\pdfoutput=1

\documentclass[11pt]{article}

\usepackage{naacl2021}

\usepackage{times}
\usepackage{latexsym}
\usepackage{verbatim}
\usepackage{booktabs}

\usepackage[T1]{fontenc}
\usepackage[utf8]{inputenc}

\usepackage{microtype}
\usepackage{subcaption}
\usepackage{mathtools}
\usepackage{multirow}
\usepackage{amsmath,amsfonts,bm}
\usepackage{amssymb}%
\usepackage{bbm}
\usepackage{algorithm}
\usepackage{algpseudocode}
\usepackage{algcompatible}
\usepackage{todonotes}
\usepackage{xspace}

\usepackage{hyperref}

\usepackage{url}
\usepackage{todonotes}

\newcommand{\sysname}{{\mbox{ASTRA}}\xspace}
\definecolor{forestgreen}{rgb}{0.13, 0.55, 0.13}

\newcommand{\gz}[1]{}%
\newcommand{\gk}[1]{}%
\title{Self-Training with Weak Supervision}
\author{Giannis Karamanolakis$^\S$ \thanks{\hspace{1.5mm}Most of the work was done while the first author was an intern at Microsoft Research.} \quad
 Subhabrata Mukherjee$^\dag$ \\ 
 {\bf Guoqing Zheng$^\dag$ \quad
 Ahmed Hassan Awadallah$^\dag$}
\\
  $^\S$Columbia University, New York  \qquad$^\dag$Microsoft Research  \\ 
  gkaraman@cs.columbia.edu \\
  \{submukhe, zheng, hassanam\}@microsoft.com
}

\begin{document}
\maketitle
\begin{abstract}
State-of-the-art deep neural networks require large-scale labeled training data that is often expensive to obtain or not available for many tasks. 
Weak supervision in the form of domain-specific rules has been shown to be useful in such settings to automatically generate weakly labeled training data. 
However, learning with weak rules is challenging due to their inherent heuristic and noisy nature. An additional challenge is rule coverage and overlap, where prior work on weak supervision only considers instances that are covered by weak rules, thus leaving valuable unlabeled data behind. 

In this work, we develop a weak supervision framework (\sysname\footnote{\sysname: we{\bf A}kly-supervised {\bf S}elf-{\bf TRA}ining. Our code is publicly available at \url{https://github.com/microsoft/ASTRA}.}) that leverages all the available data for a given task. To this end, we leverage task-specific unlabeled data through self-training with a model (student) that considers contextualized representations and predicts pseudo-labels for instances that may not be covered by weak rules. 
We further develop a rule attention network (teacher) that learns how to aggregate student pseudo-labels with weak rule labels, conditioned on their fidelity and the underlying context of an instance. 
Finally, we construct a semi-supervised learning objective for end-to-end training with unlabeled data, domain-specific rules, and a small amount of labeled data.
Extensive experiments on six benchmark datasets for text classification demonstrate the effectiveness of our approach with significant improvements over state-of-the-art baselines.

\end{abstract}

\section{Introduction}
The success of state-of-the-art neural networks crucially hinges on the availability of large amounts of annotated training data. While recent advances on language model pre-training~\cite{peters2018deep,devlin2019bert,radford2019language} reduce the annotation bottleneck, they still require large amounts of labeled data for obtaining state-of-the-art performances on downstream tasks. %
However, it is prohibitively expensive to obtain large-scale labeled data for every new task, therefore posing a significant challenge for supervised learning.

\begin{figure}
    \centering
    \includegraphics[width=\columnwidth]{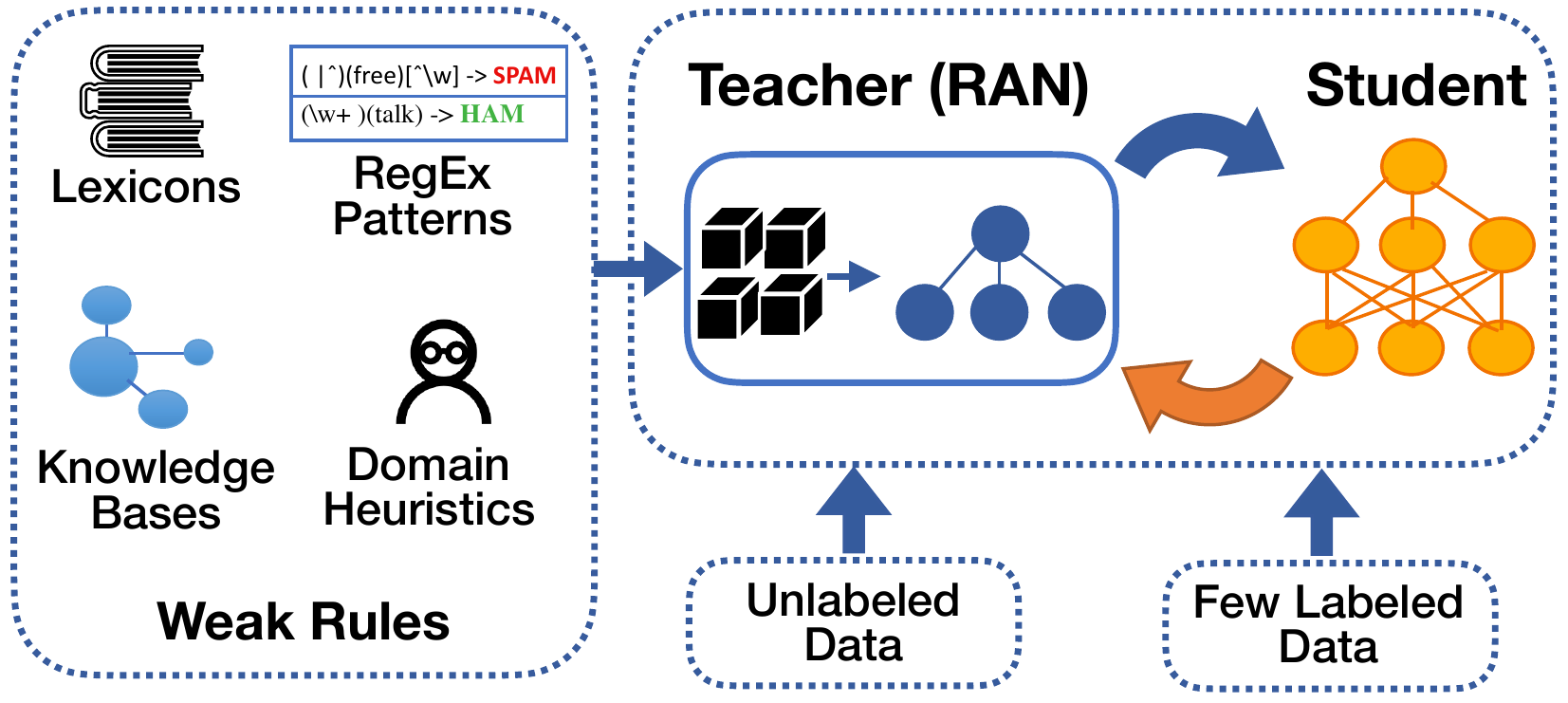}
    \caption{Our weak supervision framework, \sysname, leverages domain-specific rules, a large amount of (task-specific) unlabeled data, and a small amount of labeled data via iterative self-training.%
    }
    \label{fig:intro_architecture}
\end{figure}

In order to mitigate labeled data scarcity, recent works have tapped into weak or noisy sources of supervision, 
such as regular expression patterns~\cite{augenstein2016stance}, class-indicative keywords~\cite{ren2018learning,karamanolakis2019leveraging}, alignment rules over existing knowledge bases~\cite{mintz2009distant,xu2013filling} or heuristic labeling functions~\cite{ratner2017snorkel,bach2019snorkel,badene2019data,awasthi2019learning}.
These different types of sources can be used as weak rules for heuristically annotating large amounts of unlabeled data. 
For instance, consider the question type classification task from the TREC dataset with regular expression patterns such as: {\em label all questions containing the token ``when'' as {\tt numeric }} (e.g., ``When was Shakespeare born?").  
Approaches relying on such weak rules typically suffer from the following challenges. (i) {\em Noise.} Rules by their heuristic nature rely on shallow patterns and may predict wrong labels for many instances. For example, the question ``When would such a rule be justified?" refers to circumstances rather than numeric expressions.  
(ii) {\em Coverage.} Rules generally have a low coverage as they assign labels to only specific subsets of instances. 
(iii) {\em Conflicts.} Different rules may generate conflicting predictions for the same instance, making it  challenging to %
train a robust classifier.
To address the challenges with conflicting and noisy rules, existing approaches learn weights indicating how much to trust individual rules. In the absence of large-scale manual annotations, the rule weights are usually learned via mutual agreement and disagreement of rules over unlabeled data~\cite{ratner2017snorkel,platanios2017estimating,sachan2018learning,bach2019snorkel, ratner2019training,awasthi2019learning}. For instance, such techniques would up-weight rules that agree with each other (as they are more likely to be correct), and down-weight such rules otherwise. 
An important drawback of these approaches is low coverage since rules assign weak labels to only a subset of the data, thus leading to low rule overlap to compute rule agreement. For instance, in our experiments on six real-world datasets, we observe that $66\%$ of the instances are covered by fewer than $2$ rules and $40\%$ of the instances are not covered by any rule at all. 
Rule sparsity limits the effectiveness of previous approaches, thus leading to strong assumptions, such as, that each rule has the same weight across all instances~\cite{ratner2017snorkel, bach2019snorkel, ratner2019training}, or that additional supervision is available in the form of labeled ``exemplars'' used to create such rules in the first place~\cite{awasthi2019learning}.  
Most importantly, all these works ignore (as a data pre-processing step) unlabeled instances that are not covered by any of the rules, thus leaving potentially valuable data behind.

\noindent{\bf Overview of our method.} In this work, we present a weak supervision framework, namely \sysname, that considers {all} task-specific unlabeled instances and domain-specific rules without strong assumptions about the nature or source of the rules. \sysname makes effective use of a small amount of labeled data, lots of task-specific unlabeled data, and domain-specific rules through iterative teacher-student co-training (see Figure~\ref{fig:intro_architecture}). A student model based on contextualized representations provides pseudo-labels for all instances, thereby, allowing us to leverage all unlabeled data including instances that are not covered by any heuristic rules. To deal with the noisy nature of heuristic rules and pseudo-labels from the student, we develop a rule attention (teacher) network that learns to predict the fidelity of these rules and pseudo-labels conditioned on the context of the instances to which they apply. We develop a semi-supervised learning objective based on minimum entropy regularization to learn all of the above tasks jointly without the requirement of additional rule-exemplar supervision.

\begin{figure*}
    \centering
    \includegraphics[width=2\columnwidth]{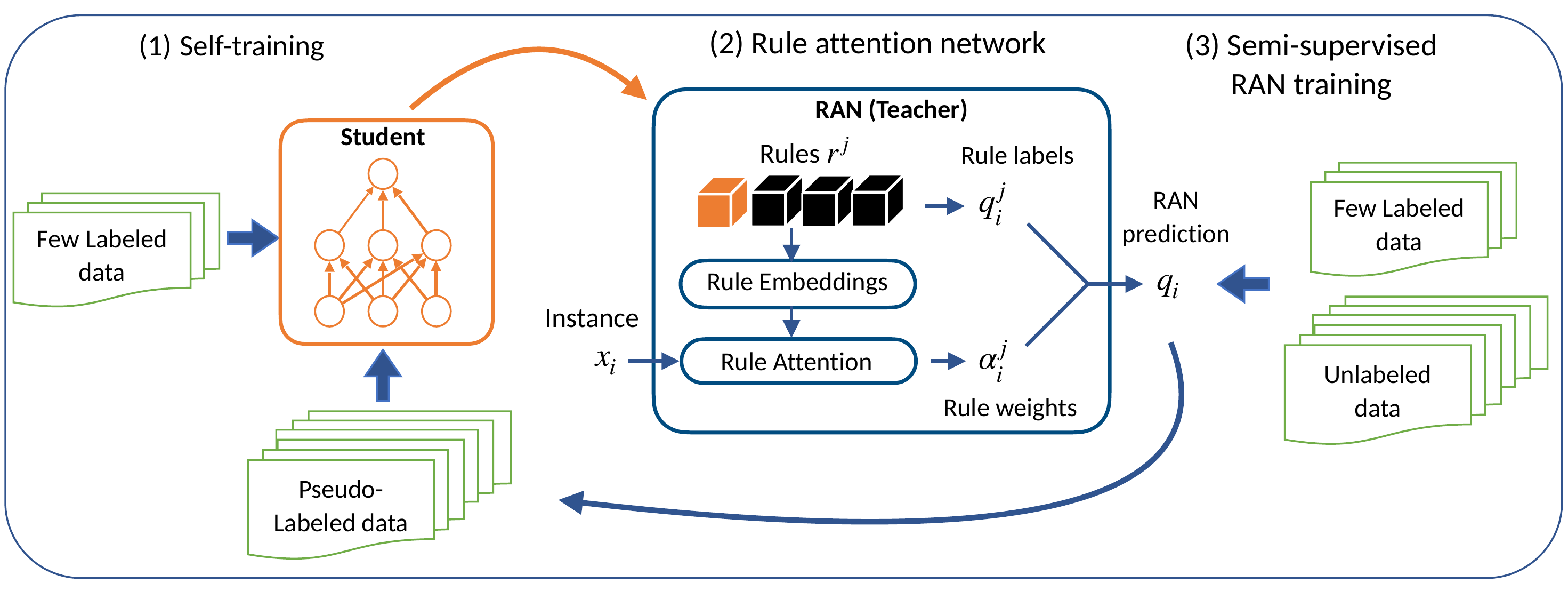}
    \caption{Our \sysname framework for self-training with weak supervision.}
    \label{fig:system_overview}
\end{figure*}

Overall, we make the following contributions: 
\begin{itemize}
    \item We propose an iterative self-training mechanism for training deep neural networks with weak supervision by making effective use of task-specific unlabeled data and domain-specific heuristic rules. The self-trained student model predictions augment the weak supervision framework with instances that are not covered by rules.
    \item We propose a rule attention teacher network (RAN) for combining multiple rules and student model predictions with instance-specific weights conditioned on the corresponding contexts. Furthermore, we construct a semi-supervised learning objective for training RAN without strong assumptions about the structure or nature of the weak rules. 
    \item We demonstrate the effectiveness of our approach on several benchmark datasets for text classification where our method significantly outperforms state-of-the-art weak supervision methods. %
\end{itemize}

\section{Self-Training with Weak Supervision}
We now present our approach, \sysname, that leverages a small amount of labeled data, a large amount of unlabeled data, and domain-specific heuristic rules.
Our architecture has two main components: the base student model (Section~\ref{ss:base-model}) and the rule attention teacher network (Section~\ref{ss:ran}), which are iteratively co-trained in a self-training framework. 
Formally, let $\mathcal{X}$ denote the instance space and $\mathcal{Y} = \{1, \dots , K\}$ denote the label space for a $K$-class classification task. 
We consider a small set of manually-labeled examples $D_L = \{(x_l, y_l)\}$, where $x_l \in \mathcal{X}$ and $y_l \in \mathcal{Y}$ and a large set of unlabeled examples $D_U = \{x_i\}$.
We also consider a set of pre-defined heuristic rules $R = \{r^j\}$, where each rule $r^j$ has the general form of a labeling function that considers as input an instance $x_i \in \mathcal{X}$ (and potentially additional side information), and either assigns a {\em weak} label $q_i^j \in {\{0, 1\}}^K$ (one-hot encoding) or does not apply, i.e., does not assign a label for $x_i$.
Our goal is to leverage $D_L$, $D_U$, and $R$ to train a classifier that, given an unseen test instance $x' \in \mathcal{X}$, predicts a label $y'\in \mathcal{Y}$.
In the rest of this section, we present our \sysname framework for addressing this problem.

\subsection{Base Student Model}
\label{ss:base-model}
Our self-training framework starts with a base model trained on the available small labeled set $D_L$. The model is then applied to unlabeled data $D_U$ to obtain pseudo-labeled instances. 
In classic self-training~\cite{riloff1996automatically,nigam2000analyzing}, the student model's pseudo-labeled instances are directly used to augment the training dataset and iteratively re-train the student. 
In our setting, we augment the self-training process with weak labels drawn from our teacher model that also considers rules in $R$ (described in the next section).
The overall self-training process can be formulated as:
%
\begin{multline}
\label{eq:st}
    \min_\theta\ \mathbb{E}_{x_l, y_l \in D_L} [-\log\ p_\theta (y_l \mid x_l)] +\\ \lambda \mathbb{E}_{x \in D_U} \mathbb{E}_{y \sim q_{\phi^*}(y \mid x)} [-\log\ p_\theta (y \mid x)]
\end{multline}
where, $p_\theta(y|x)$ is the conditional distribution under student's parameters $\theta$; $\lambda \in \mathbb{R}$ is a hyper-parameter controlling the relative importance of the two terms; and $q_{\phi^*}(y\mid x)$ is the conditional distribution under the teacher's parameters $\phi^*$ from the last iteration that is fixed in the current iteration. 

\subsection{Rule Attention Teacher Network (RAN)}
\label{ss:ran}
Our Rule Attention Teacher Network (RAN) aggregates multiple weak sources of supervision with trainable weights and computes a soft weak label $q_i$ for an unlabeled instance $x_i$. 
One of the potential drawbacks of relying only on heuristic rules is that a lot of data get left behind. Heuristic rules by nature (e.g., regular expression patterns, keywords) apply to only a subset of the data. Therefore, a substantial number of instances are not covered by any rules and thus are not considered in prior weakly supervised learning approaches~\citep{ratner2017snorkel,awasthi2019learning}. To address this challenge and leverage contextual information from all available task-specific unlabeled data, we leverage the corresponding pseudo-labels predicted by the base student model (from Section~\ref{ss:base-model}). To this end, we apply the student to the unlabeled data $x \in D_U$ and obtain pseudo-label predictions as $p_\theta(y|x)$. These predictions are used to augment the set of already available weak rule labels to increase rule coverage.

Let $R_i \subset R$ be the set of all heuristic rules that {\em apply} to instance $x_i$.
The objective of RAN is to aggregate the weak labels predicted by all rules $r^j \in R_i$ and the student pseudo-label $p_\theta(y|x_i)$ to compute a soft label $q_i$ for every instance $x_i$ from the unlabeled set $D_U$. %
In other words, RAN considers the student as an additional source of weak rule.
Aggregating all rule labels into a single label $q_i$ via simple majority voting (i.e., predicting the label assigned by the majority of rules) may not be effective as it treats all rules equally, while in practice, certain rules are more accurate than others.

RAN predicts pseudo-labels $q_i$ by aggregating rules with trainable weights $a^{(\cdot)}_i \in [0,1]$ that capture their fidelity towards an instance $x_i$ as:
%
\begin{equation}
    q_i = \frac{1}{Z_i} \bigg( \sum_{j: \ r^j \in R_i} a^j_i q^j_i + a_i^S  p_\theta(y|x_i) +  a^u_i u \bigg),
\label{eq:ran-aggregation}
\end{equation}
where $a_i^j$ and $a_i^S$ are the fidelity weights for the heuristic rule labels $q_i^j$ and the student assigned pseudo-label $p_\theta(y|x_i)$ for an instance $x_i$, respectively; $u$ is a uniform rule distribution that assigns equal probabilities for all the $K$ classes as $u=[\frac{1}{K}, \dots, \frac{1}{K}]$; 
$a^u_i$ is the weight assigned to the ``uniform rule'' for $x_i$, which is computed as a function of the rest of the rule weights:
$a^u_i = (|R_i| + 1 -\sum_{j: \ r^j \in R_i} a^j_i  -a^{S}_i)$;
and $Z_i$ is a normalization coefficient to ensure that $q_i$ is a valid probability distribution. %
$u$ acts as a uniform smoothing factor that prevents overfitting for sparse settings, for instance, when a single weak rule applies to an instance.

According to Eq.~\eqref{eq:ran-aggregation}, a rule $r^j$ with higher fidelity weight $a^j_i$ contributes more to the computation of $q_i$.  
If $a^j_i = 1 \  \forall r^j \in \{R_i \cup p_\theta\}$, then RAN reduces to majority voting.
If $a^j_i = 0 \  \forall r^j \in \{R_i \cup p_\theta\}$, then RAN ignores all rules and predicts $q_i=u$. 
Note the distinction of our setting to recent works like Snorkel~\cite{ratner2017snorkel}, that learns global rule-weights $a^j_i = a^j \ \forall x_i$ by ignoring the instance-specific rule fidelity.
Our proposed approach is flexible but at the same time challenging as we do not assume prior knowledge of the internal structure of the labeling functions $r^j \in R$. 

In order to effectively compute rule fidelities, RAN considers instance embeddings that capture the context of instances beyond the shallow patterns considered by rules. 
In particular, we model the weight $a^j_i$ of rule $r^j$ as a function of the context of the instance $x_i$ and $r^j$ through an attention-based mechanism. 
Consider $h_i \in \mathbb{R}^{d'}$ to be the hidden state representation of $x_i$ from the base student model. 
Also, consider the (trainable) embedding of each rule $r^j$ as $e_j = g(r^j) \in \mathbb{R}^d$. 
We use $e_j$ as a query vector with {\em sigmoid attention} to compute instance-specific rule attention weights as: 
\begin{equation}
    a_i^j = \sigma(f(h_i)^T \cdot e_j) \in [0,1],
    \label{eq:rule-weight-formula}
\end{equation}
where $f$ is a multi-layer perceptron that projects $h_i$ to $\mathbb{R}^d$ and $\sigma(\cdot)$ is the sigmoid function. 
Rule embedding allows us to exploit the similarity between different rules in terms of instances to which they apply, and further leverage their semantics for modeling agreement. 
RAN computes the student's weight $a_i^S$ using the same procedure as for computing the rule weights $a_i^j$.

Note that the rule predictions $q^j_i$ are considered fixed, while we estimate their attention weights. 
The above coupling between rules and instances via their corresponding embeddings $e_j$ and $h_i$ allows us to obtain representations where similar rules apply to similar contexts, and model their agreements via the attention weights $a_i^j$. 
To this end, the trainable parameters of RAN ($f$ and $g$) are shared across all rules and instances. 
Next, we describe how to train RAN. 

\begin{figure}
    \centering
    \begin{subfigure}[t]{0.49\columnwidth}
        \centering
        \includegraphics[width=0.99\columnwidth]{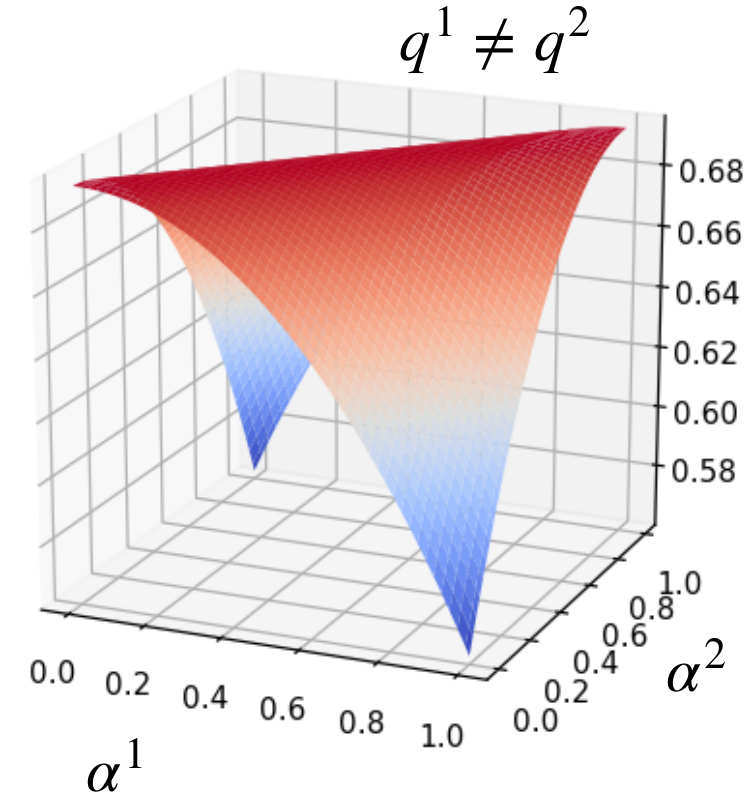}
        \caption{Rule predictions disagree.}
    \label{fig:loss_agree}
    \end{subfigure}%
    ~ 
    \begin{subfigure}[t]{0.49\columnwidth}
        \centering
        \includegraphics[width=0.95\columnwidth]{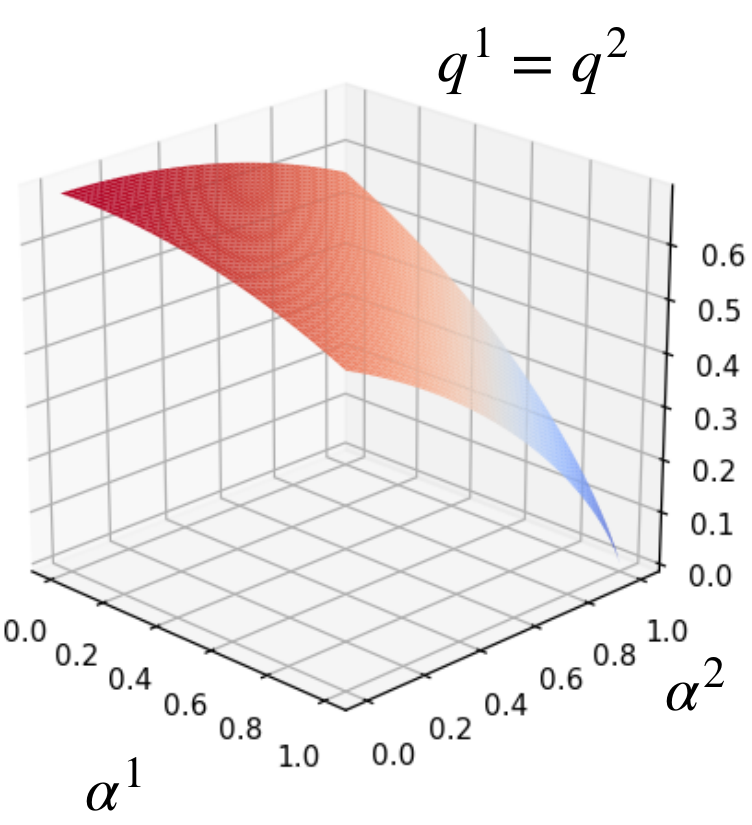}
        \caption{Rule predictions agree.}
        \label{fig:loss_disagree}
    \end{subfigure}%
    \caption{Variation in unsupervised entropy loss with instance-specific rule predictions and attention weights encouraging rule agreement. Consider this illustration with two rules for a given instance. When rule predictions disagree ($q^1$ $\neq$ $q^2$), minimum loss is achieved for attention weights $a^1$=0, $a^2$=1 or $a^1$=1, $a^2$=0. When rule predictions agree ($q^1$=$q^2$), minimum loss is achieved for attention weights $a^1$=$a^2$=1. For instances covered by three rules, if $q^1$=$q^2$$\neq$$q^3$, the minimum loss is achieved for $a^1$=$a^2$=1 and $a^3$=0.}
    \label{fig:unsup_loss1}
\end{figure}

\subsection{Semi-Supervised Learning of \sysname}
\label{ss:ssl-ran-training}
Learning to predict instance-specific weights $a_i^{(\cdot)}$ for the weak sources (including rules and student pseudo-labels) is challenging due to the absence of any explicit knowledge about the source quality and limited amount of labeled training data. We thus treat the weights $a_i^{(\cdot)}$ as latent variables and propose a semi-supervised objective for training RAN with supervision on the coarser level of $q_i$:
%
%
%
\begin{equation}
    \mathcal{L}^{RAN} = -\sum_{(x_i, y_i) \in D_L} y_i \log q_i  - \sum_{x_i \in D_U} q_i \log q_i.
    \label{eq:ran-ssl-objective}
\end{equation}
Given task-specific labeled data $D_L$, the first term in Eq.~\eqref{eq:ran-ssl-objective} minimizes the cross-entropy loss between the teacher's label $q_i$ and the corresponding clean label $y_i$ for the instance $x_i$. This term penalizes weak sources that assign labels $q^{(\cdot)}_i$ that contradict with the ground-truth label $y_i$ by assigning a low instance-specific fidelity weight $a^{(\cdot)}_i$. 

\begin{figure}[t]
    \centering
\begin{algorithm}[H]
\caption{Self-training with Weak Supervision}
\label{algo:wst}
\begin{algorithmic}[1]
\Statex{\textbf{Input:} Small amount of labeled data $D_L$; task-specific unlabeled data $D_U$; weak rules $R$}
\Statex{\textbf{Outputs:}} Student $p_\theta^*(\cdot)$, RAN Teacher $q_\phi^*(\cdot) $%
    \State{Train student $p_\theta(\cdot)$ using $D_L$}
        \State{\textbf{Repeat until convergence:}}
      \begin{algorithmic}
      \State{{\small2.1:} Train teacher $q_\phi(\cdot)$ using $D_L$, $D_U$ through Eq.~\eqref{eq:ran-aggregation} and~\eqref{eq:ran-ssl-objective}}
        \State{{\small2.2:} Apply $q_\phi(y\mid x, R, p_\theta)$ to $x \in D_U$ to obtain pseudo-labeled data: $D_{RAN}=\{(x_i, q_i)\}_{x_i \in D_U}$ through Eq.~\eqref{eq:ran-aggregation}}
        \State{{\small2.3:} Train $p_\theta(\cdot)$ using $D_L$, $D_{RAN}$ through Eq.~\eqref{eq:st}} 
     \end{algorithmic}
\end{algorithmic}
\end{algorithm}
\end{figure}

\begin{table*}[!htb]
\resizebox{2\columnwidth}{!}{
\begin{tabular}{llllllll}
\toprule 
& \textbf{TREC}  & \textbf{SMS} &\textbf{YouTube} &  \textbf{CENSUS} & \textbf{MIT-R}  & \textbf{Spouse}  \\\midrule 
Labeled Training Data ($|D_L|$) & 68      & 69 & 100      & 83   & 1842     &  100 \\
Unlabeled Training Data ($|D_U|$) & 5K       & 5K  & 2K  & 10K   & 65K       & 22K \\
Test Data & 500  & 500  & 250   & 16K   & 14K      & 3K \\
\#Classes & 6    & 2     & 2      & 2 & 9  & 2 \\
\#Rules & 68       & 73   & 10   & 83   & 15     & 9 \\
Rule Accuracy (Majority Voting) & 60.9\% & 48.4\%& 82.2\% & 80.1\%& 40.9\% &  44.2\%\\
Rule Coverage (instances in $D_U$ covered by $\geq 1$ rule)& 95\% & 40\%& 87\%  & 100\% & 14\%   & 25\%\\
Rule Overlap (instances in $D_U$ covered by $\geq 2$ rules)& 46\%  & 9\% & 48\%  & 94\%& 1\%   & 8\%\\ 
\bottomrule   
\end{tabular}}
\caption{Dataset statistics. }
\label{tab:dataset-statistics}
\end{table*}

The second term in Eq.~\eqref{eq:ran-ssl-objective} 
minimizes the entropy of the aggregated pseudo-label $q_i$ on unlabeled data $D_U$.
Minimum entropy regularization is effective in settings with small amounts of labeled data by leveraging unlabeled data~\citep{grandvalet2005semi}, and  is highly beneficial in our setting because it encourages RAN to predict weights that maximize rule agreement. 
Since the teacher label $q_i$ is obtained by aggregating weak labels $q_i^{(\cdot)}$, entropy minimization encourages RAN to predict higher instance-specific weights $a^{(\cdot)}_i$ to sources that agree in their labels for $x_i$, and lower weights when there are disagreements between weak sources -- aggregated across all the unlabeled instances.

Figure~\ref{fig:unsup_loss1} plots the minimum entropy loss over unlabeled data over two scenarios where two rules agree or disagree with each other for a given instance. The optimal instance-specific fidelity weights $a_i^{(\cdot)}$ are $1$ when rules agree with each other, thereby, assigning credits to both rules, and only one of them when they disagree. We use this unsupervised entropy loss in conjunction with cross-entropy loss over labeled data to ensure grounding.

\noindent{\bf End-to-end Learning:}  Algorithm~\ref{algo:wst} presents an overview of our learning mechanism. We first use the small amount of labeled data to train a base student model that generates pseudo-labels and augments heuristic rules over unlabeled data. Our RAN network computes fidelity weights to combine these different weak labels via minimum entropy regularization to obtain an aggregated pseudo-label for every unlabeled instance. This is used to re-train the student model with the above student-teacher training repeated till convergence.
\section{Experiments}
\paragraph{Datasets.} We evaluate our framework on the following six benchmark datasets for weak supervision from ~\citet{ratner2017snorkel} and~\citet{awasthi2019learning}. (1) Question classification from TREC-6 into $6$ categories (Abbreviation, Entity, Description, Human, Location, Numeric-value); (2) Spam classification of SMS messages; (3) Spam classification of YouTube comments; (4) Income classification on the CENSUS dataset on whether a person earns more than \$50K or not; (5) Slot-filling in sentences on restaurant search queries in the MIT-R dataset: each token is classified into $9$ classes (Location, Hours, Amenity, Price, Cuisine, Dish, Restaurant Name, Rating, Other); (6) Relation classification in the Spouse dataset, whether pairs of people mentioned in a sentence are/were married or not. 

Table~\ref{tab:dataset-statistics} shows the dataset statistics along with the amount of labeled, unlabeled data and domain-specific rules for each dataset. 
For a fair comparison, we use exactly the same set of rules as in the previous work for the benchmark datasets. These rules include regular expression patterns, lexicons, and knowledge bases for weak supervision. 
Most of these rules were constructed manually, except for the CENSUS dataset, where rules have been automatically extracted with a coverage of 100\%. %

On average across all the datasets, $66\%$ of the instances are covered by fewer than 2 rules, whereas $40\%$ are not covered by any rule at all -- demonstrating the sparsity in our setting. We also report the accuracy of the rules in terms of majority voting on the task-specific unlabeled datasets.
Additional details on the dataset and examples of rules are presented in the Appendix.

\begin{table}[]
    \centering
    \resizebox{\columnwidth}{!}{
    \begin{tabular}{lccc}
    \toprule 
        \multirow{3}{*}{\textbf{Method}} & \multicolumn{2}{c}{\textbf{Learning to Weight}} &  \textbf{Unlabeled}\\
         & \textbf{Rules} & \textbf{Instances} & \textbf{(no rules)}\\\midrule  
        Majority & - & - & -\\
        Snorkel{\small ~\cite{ratner2017snorkel}} & \checkmark & - & -\\
     PosteriorReg{\small ~\cite{hu2016harnessing}} & \checkmark & - & -\\
    L2R{\small ~\cite{ren2018learningb}} & - & \checkmark &  -\\
ImplyLoss{\small ~\cite{awasthi2019learning}} & \checkmark & \checkmark &  -\\
    Self-train & - & - &  \checkmark\\
\sysname & \checkmark & \checkmark &  \checkmark\\
    \bottomrule 
    \end{tabular}}
    \caption{\sysname learns rule-specific and instance-specific attention weights and leverages task-specific unlabeled data where no rules apply.}
    \label{tab:method_comparison}
\end{table}

\begin{table*}[]
\centering
\resizebox{1.7\columnwidth}{!}{
\begin{tabular}{lcccccc}
\toprule
  & \textbf{TREC}  & \textbf{SMS}   & \textbf{YouTube} & \textbf{CENSUS}  & \textbf{MIT-R}  & \textbf{Spouse} \\
                      & \textbf{ (Acc)} & \textbf{(F1)}   & \textbf{(Acc)} & \textbf{(Acc)} &  \textbf{(F1)} & \textbf{(F1)}  \\\midrule 

Majority            & 60.9 {\small(0.7)} & 48.4 {\small(1.2)} & 82.2 {\small(0.9)}    & 80.1 {\small(0.1)}   & 40.9 {\small(0.1)} & 44.2 {\small(0.6)}   \\
LabeledOnly       & 66.5 {\small(3.7)} & 93.3 {\small(2.9)} & 91.0 {\small(0.7)}    & 75.8 {\small(1.7)}   & 74.7 {\small(1.1)} & 47.9 {\small(0.9)}  \\\midrule 
Snorkel+Labeled  & 65.3 {\small(4.1)}& 94.7 {\small(1.2)} & 93.5 {\small(0.2)}    & 79.1 {\small(1.3)}   & 75.6 {\small(1.3)} & 49.2 {\small(0.6)}   \\
PosteriorReg   & 67.3 {\small(2.9)} & 94.1 {\small(2.1)} & 86.4 {\small(3.4)}    & 79.4 {\small(1.5)}   & 74.7 {\small(1.2)} & 49.4 {\small(1.1)}      \\
L2R     & 71.7 {\small(1.3)} & 93.4 {\small(1.1)} & 92.6 {\small(0.5)}    & 82.4 {\small(0.1)}   & 58.6 {\small(0.4)} & 49.5 {\small(0.7)}  \\
ImplyLoss          & 75.5 {\small(4.5)} & 92.2 {\small(2.1)} & 93.6 {\small(0.5)}    & 80.5 {\small(0.9)}   & 75.7 {\small(1.5)} & 49.8 {\small(1.7)}   \\
Self-train & 71.1 {\small(3.9)} & 95.1 {\small(0.8)} & 92.5 {\small(3.0)}    & 78.6 {\small(1.0)}   & 72.3 {\small(0.6)} &51.4 {\small(0.4)}   \\
\sysname (ours) & \textbf{80.3} {\small(2.4)} & \textbf{95.3} {\small(0.5)} & \textbf{95.3} {\small(0.8)}    & \textbf{83.1} {\small(0.4)}   & \textbf{76.9} {\small(0.6)} & \textbf{62.3} {\small(1.1)}  \\\bottomrule
\end{tabular}
}
\caption{Overall result comparison across multiple datasets. Results are aggregated over five runs with random training splits and standard deviation across the runs in parentheses.} %
\label{tab:main-results}
\end{table*}

\noindent{\bf Evaluation.} We train \sysname five times for five different random splits of the labeled training data and evaluate on held-out test data. We report the average performance as well as the standard deviation across multiple runs. We report the same evaluation metrics as used in prior works~\cite{ratner2017snorkel,awasthi2019learning} for a fair comparison.

\noindent{\bf Model configuration.} 
Our student model consists of embeddings from pre-trained language models like ELMO~\cite{peters2018deep} or BERT~\cite{devlin2019bert} for generating contextualized representations for an instance, followed by a softmax classification layer.
The RAN teacher model considers a rule embedding layer and a multilayer perceptron for mapping the contextualized representation for an instance to the rule embedding space. Refer to the Appendix for more details.
\noindent{\bf Baselines.} We compare our method with the following methods: \emph{(a) Majority} predicts the majority vote of the rules with ties resolved by predicting a random class.
\emph{(b) LabeledOnly} trains classifiers using only labeled data (fully supervised baseline).
\emph{(c) Self-train}~\cite{nigam2000analyzing,lee2013pseudo} leverages both labeled and unlabeled data for iterative self-training on pseudo-labeled predictions over task-specific unlabeled data. This baseline ignores domain-specific rules. 
\emph{(e) Snorkel+Labeled}~\cite{ratner2017snorkel} trains classifiers using weakly-labeled data with a generative model. The model is trained on unlabeled data for computing rule weights in an unsupervised fashion, and learns a single weight per rule across all instances. It is further fine-tuned on labeled data. 
\emph{(f) L2R}~\cite{ren2018learning} learns to re-weight noisy or weak labels from domain-specific rules via meta-learning. It learns instance-specific but not rule-specific weights. %
\emph{(g) PosteriorReg}~\cite{hu2016harnessing} trains classifiers using rules as soft constraints via posterior regularization~\cite{ganchev2010posterior}. %
\emph{(h) ImplyLoss}~\cite{awasthi2019learning} leverages {\em exemplar}-based supervision as additional knowledge for learning instance-specific and rule-specific weights by minimizing an implication loss over unlabeled data. This requires maintaining a record of all instances used to create the weak rules in the first place. %
Table~\ref{tab:method_comparison} shows a summary of the different methods contrasting them on how they learn the weights (rule-specific or instance-specific) and if they leverage task-specific unlabeled data that are not covered by any rules.

\subsection{Experimental Results}

\noindent{\bf Overall results.} Table~\ref{tab:main-results} summarizes the main results across all datasets. %
Among all the semi-supervised methods that leverage weak supervision from domain-specific rules, \sysname outperforms Snorkel by $6.1\%$ in average accuracy across all datasets by learning instance-specific rule weights in conjunction with self-training over unlabeled instances where weak rules do not apply.  %
Similarly, \sysname also improves over a recent work and the best performing baseline ImplyLoss by $3.1\%$ on average. Notably, our method does not require additional supervision at the level of exemplars used to create rules in contrast to ImplyLoss. %

\noindent{\bf Self-training over unlabeled data.} Recent works for tasks like image classification~\cite{li2019learning,xie2020self,zoph2020rethinking}, neural sequence generation~\cite{zhang2016exploiting,he2019revisiting} and few-shot text classification~\cite{mukherjee2020uncertainty, wang2021adaptive} show the effectiveness of self-training methods in exploiting task-specific unlabeled data with stochastic regularization techniques like dropouts and data augmentation. We also make similar observations for our weakly supervised tasks, where classic self-train methods (``Self-train'') leveraging only a few task-specific labeled examples and lots of unlabeled data outperform weakly supervised methods like Snorkel and PosteriorReg that have additional access to domain-specific rules.

\noindent{\bf Self-training with weak supervision.} Our framework \sysname provides an efficient method to incorporate weak supervision from domain-specific rules to augment the self-training framework and improves by $6\%$ over classic self-training. 

\begin{figure}
    \centering
    \includegraphics[width=\columnwidth]{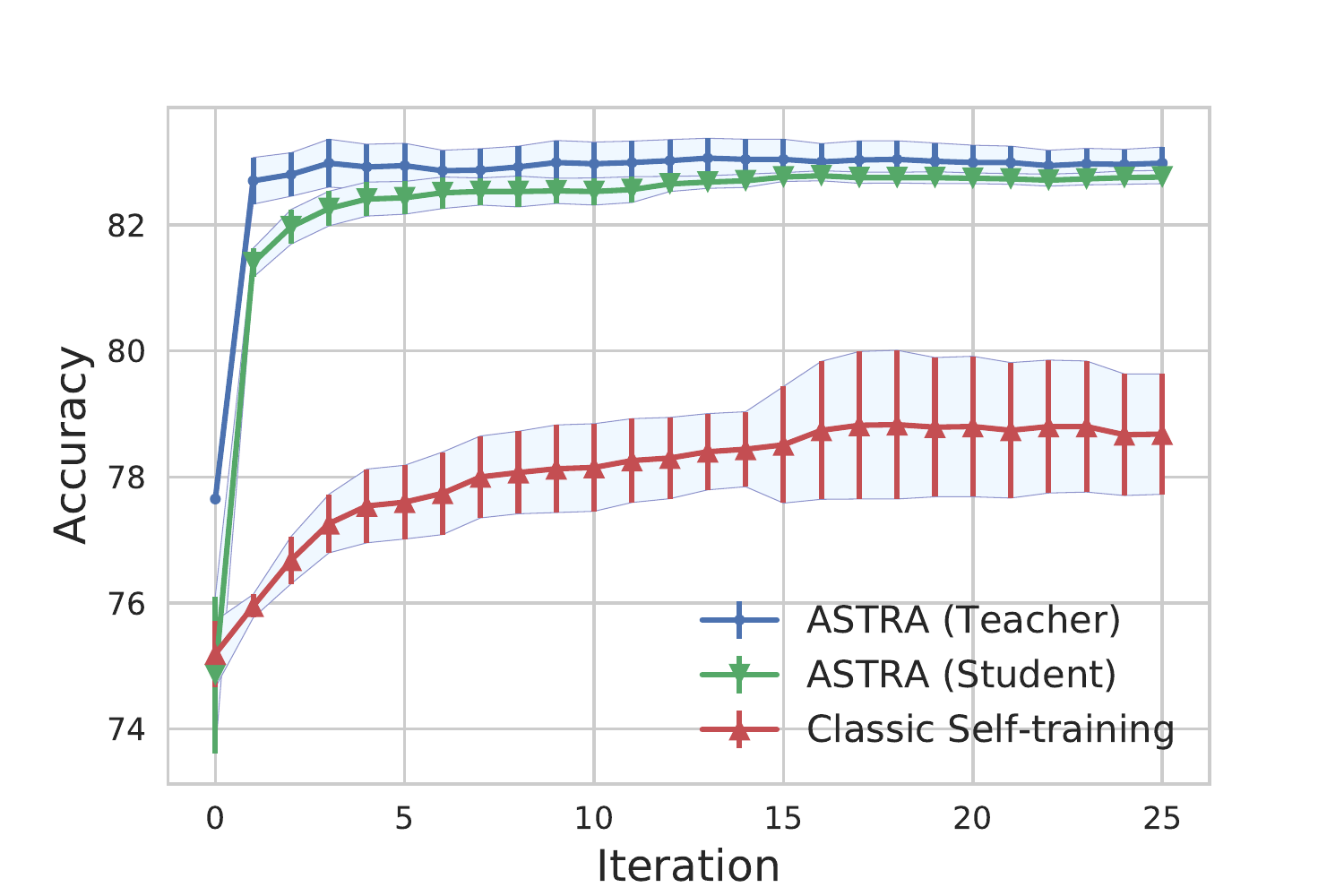}
    \caption{Gradual accuracy improvement over self-training iterations in the CENSUS dataset. \sysname (Student) performs better than Classic Self-training (Student) being guided by a better teacher. %
    }
    \label{fig:selftrain_iteration}
\end{figure}

To better understand the benefits of our approach compared to classic self-training, consider Figure~\ref{fig:selftrain_iteration}, which depicts the gradual performance improvement over iterations. The student models in classic self-training and \sysname have exactly the same architecture. However, the latter is guided by a better teacher (RAN) that learns to aggregate noisy rules and pseudo-labels over unlabeled data.

\begin{figure}
    \centering
    \includegraphics[width=\columnwidth]{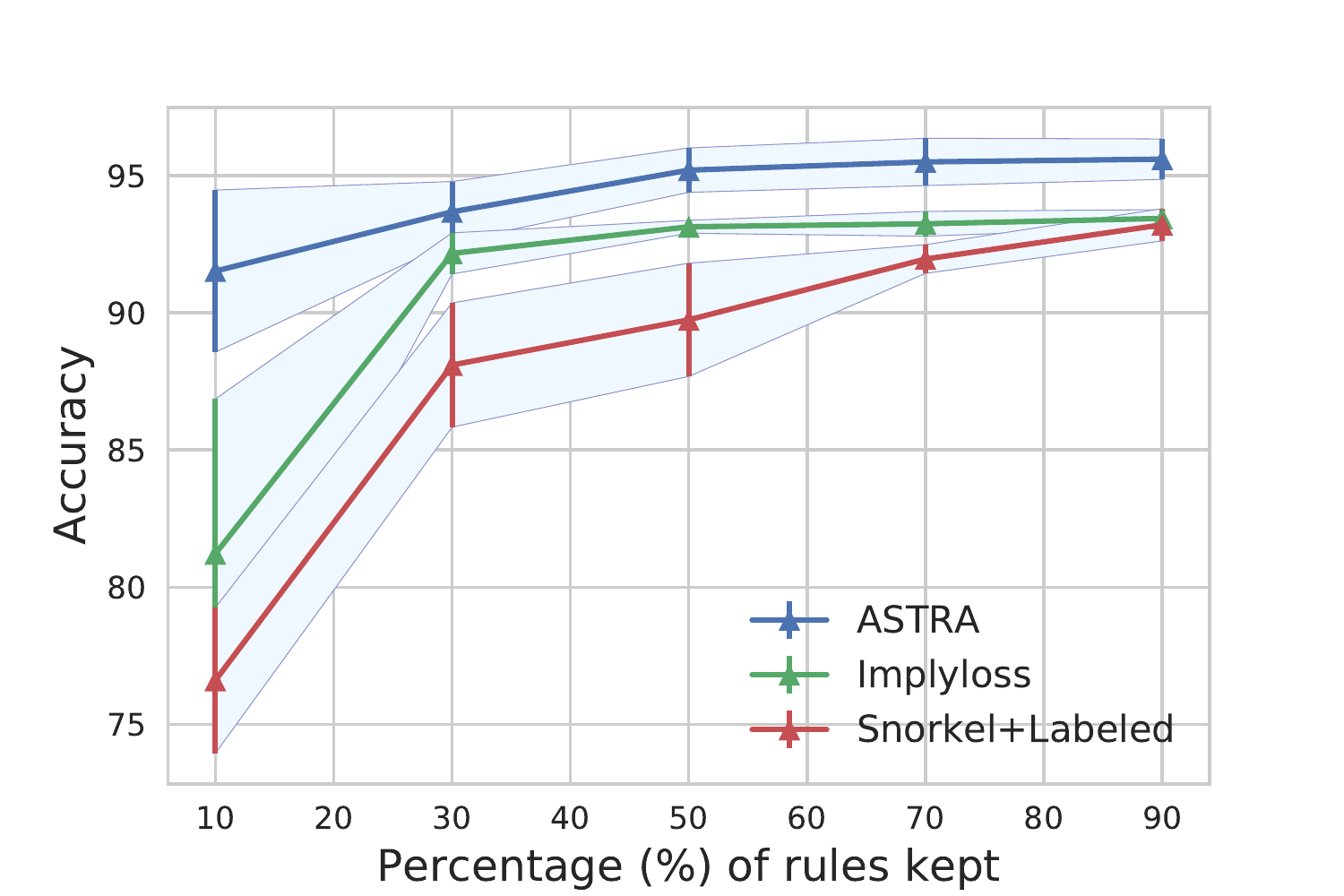}
    \caption{Performance improvement on increasing the proportion of weak rules in YouTube. For each setting, we randomly sample a subset of rules, aggregate and report results across multiple runs. \sysname is effective across all settings with strongest improvements under high rule sparsity (left region of the x-axis).}
    \label{fig:rule_sparsity_curve}
\end{figure}

\noindent{\bf Impact of rule sparsity and coverage for weak supervision.}  
In this experiment, we compare the performance of various methods by varying the proportion of available domain-specific rules. To this end, we randomly choose a subset of the rules (varying the proportion from $10\%$ to $100\%$) and train various weak supervision methods. For each setting, we repeat experiments with multiple rule splits and report aggregated results in  Figure~\ref{fig:rule_sparsity_curve}. %
We observe that \sysname is effective across all settings with the most impact at high levels of rule sparsity. For instance, with $10\%$ of domain-specific rules available, \sysname outperforms ImplyLoss by $12\%$ and Snorkel+Labeled by $19\%$. 

This performance improvement is made possible by incorporating self-training in our framework to obtain pseudo-labels for task-specific unlabeled instances, and further re-weighting them with other domain-specific rules via the rule attention network. Correspondingly, Table~\ref{tab:overlap-increase} shows the increase in data coverage for every task given by the proportion of unlabeled instances that are now covered by at least two weak sources (from multiple rules and pseudo-labels) in contrast to just considering the rules.

\subsection{Ablation Study}
Table~\ref{tab:ablation_results_summary} reports ablation experiments to evaluate the impact of various components in \sysname.

\sysname teacher marginally outperforms the student model on an aggregate having access to domain-specific rules. \sysname student that is self-trained over task-specific unlabeled data and guided by an efficient teacher model significantly outperforms other state-of-the-art baselines.

\begin{table}[t]
\resizebox{\columnwidth}{!}{
\begin{tabular}{llllllll}
\toprule 
\% Overlap & \textbf{\small TREC}  & \textbf{\small YTube} & \textbf{\small SMS}   & \textbf{\small MITR} & \textbf{\small CEN.} & \textbf{\small Spouse}  \\\midrule  
Only Rules & 46 & 48  & 9 & 1   & 94 & 8\\ 
\sysname & \textbf{95}    & \textbf{87}     & \textbf{40}    & \textbf{14}    & \textbf{100}    & \textbf{25} \\\midrule
Increase & +49    & +39     & +31   & +13    & +6    & +17 \\
\bottomrule    
\end{tabular}}
\caption{\sysname substantially increases overlap (\%) determined by the proportion of unlabeled instances that are covered by at least $2$ weak sources (from multiple rules and student pseudo-labels, as applicable).}
\label{tab:overlap-increase}
\end{table}

\begin{table}[t]
    \centering
    \resizebox{\columnwidth}{!}{
    \begin{tabular}{ll}
    \toprule  
    \textbf{Configuration}     &  \textbf{Acc}\\\midrule   
    \sysname (Teacher)   &  \textbf{88.1}\\
     \sysname (Student)   &  87.7 ($\downarrow$ 0.4\%)\\
  No min. entropy regularization in Eq.~\eqref{eq:ran-ssl-objective}  &  86.9 ($\downarrow$ 1.4\%)\\
No student fine-tuning on $D_L$ (step 2.3)   &  86.7 ($\downarrow$ 1.6\%)\\
No student pseudo-labels in RAN in Eq.~\eqref{eq:ran-aggregation} &  85.3 ($\downarrow$ 3.2\%)\\
    \bottomrule
    \end{tabular}}
    \caption{Summary of ablation experiments aggregated across multiple datasets. Refer to Appendix for corresponding results in each dataset.}
    \label{tab:ablation_results_summary}
\end{table}

 \begin{table*}[!htb]
    \centering
    \small
    \begin{tabular}{llll}
    \toprule  
        Text & \multicolumn{3}{l}{\textit{What was President Lyndon Johnson 's reform program called ?}}  \\
        Clean Label & ENTY &&\\
        \sysname Teacher & \color{forestgreen}{\textbf{ENTY}} &&\\\midrule
        \textbf{Weak Source} & \textbf{Label} & \textbf{Weight}  & \textbf{Feature / Regular expression pattern}\\
        Student & \color{forestgreen}{\textbf{ENTY}} &  $a$=$1.0$ & $h_i$ (contextualized instance embedding)\\
        Rule 8 & \color{red}{\textbf{HUM}} & $a$=$1.0$ & {\small \verb!( |^)(who|what|what)[^\w] *(\w+ ){0,1}(person|!} \\
         &  &  & {\small \verb!man|woman|human|president|president)[^\w]*( |$)!} \\
        Rule 24 & \color{forestgreen}{\textbf{ENTY}} & $a$=$1.0$ & {\small \verb!( |^)(what|what)[^\w]*(\w+ ){0,1}(is|is)[^\w]*!} \\
         &  &  & {\small \verb!*([^\s]+ )*(surname|address|name|name)[^\w]*( |$)!} \\
         Rule 42 & \color{red}{\textbf{DESC}}  & $a$=$0.0$ & {\small \verb!( |^)(explain|describe|what|what)[^\w]*( |$)!} \\
         Rule 61 & \color{red}{\textbf{HUM}} & $a$=$0.0$ & {\small \verb!( |^)(called|alias|nicknamed|nicknamed)[^\w]*( |$)!} \\
    \bottomrule 
    \end{tabular}
    \caption{Snapshot of a question in TREC-6 and corresponding predictions. 
    Top: instance text, clean label, and the aggregated prediction from \sysname teacher. Bottom: several weak rules with regular expression patterns and predicted weak labels, along with the student and its pseudo-label {\small (DESC: description, ENTY: entity, NUM: number, HUM: human)}. The weights depict the fidelity computed by RAN for each weak source for this specific instance.}
    \label{tab:detailed_TREC_example}
\end{table*}
 \begin{table*}[]
     \centering
     \resizebox{2\columnwidth}{!}{
     \begin{tabular}{llll}
     \toprule
    {\bf Instance Text (Question in TREC-6)} & {\bf Teacher} & {\bf Student} & {\bf Set of Heuristic Rule Labels}\\\midrule
       {\small 1.} \textit{Which president was unmarried?} & \color{forestgreen}{\textbf{HUM}} & {\color{forestgreen}{HUM}}{\small (1)}& \{\} \\
       {\small 2.}    \textit{What is a baby turkey called?} & \color{forestgreen}{\textbf{ENTY}} & {\color{red}{DESC}}{\small (1)}& \{{\color{forestgreen}{ENTY}}{\small (1)}, {\color{red}{DESC}}{\small (0)}, {\color{red}{HUM}}{\small (0)}\} \\
    {\small 3.}     \textit{What currency do they use in Brazil?} & \color{forestgreen}{\textbf{ENTY}} & {\color{forestgreen}{ENTY}}{\small (1)}& \{{\color{red}{DESC}}{\small (0)}, {\color{red}{DESC}}{\small (0)}\} \\
   {\small 4.}     \textit{What is the percentage of water content in the human body?} & {\color{forestgreen}{\textbf{NUM}}} & {\color{red}{DESC}}{\small (0)}& \{{\color{red}{HUM}}{\small (0)}, {\color{forestgreen}{NUM}}{\small (0.2)}, {\color{red}{DESC}}{\small (0)}\} \\
        \bottomrule
     \end{tabular}}
     \caption{Snapshot of answer-type predictions for questions in TREC-6 from \sysname teacher and student along with a set of labels assigned by various weak rules {\small (DESC: description, ENTY: entity, NUM: number, HUM: human)} with corresponding attention weights (in parentheses). Correct and incorrect predictions are colored in {\color{forestgreen}{green}} and {\color{red}{red}} respectively. Detailed analysis and rule semantics are reported in the Appendix.}
     \label{tab:trec-examples}
 \end{table*}

Through minimum entropy regularization in our semi-supervised learning objective (Eq.~\eqref{eq:ran-ssl-objective}), \sysname leverages the agreement between various weak sources (including rules and pseudo-labels) over task-specific unlabeled data. Removing this component results in an accuracy drop of $1.4\%$ on an aggregate demonstrating its usefulness.

Fine-tuning the student on labeled data is important for effective self-training: ignoring $D_L$ in the step 2.3 in Algorithm 1, leads to 1.6\% lower accuracy than \sysname.

There is significant performance drop on removing the student's pseudo-labels ($p_\theta(\cdot)$) from the rule attention network in Eq.~\eqref{eq:ran-aggregation}. This significantly limits the coverage of the teacher ignoring unlabeled instances where weak rules do not apply, thereby, degrading the overall performance by $3.2\%$.

\subsection{Case Study: TREC-6 Dataset}
Table~\ref{tab:detailed_TREC_example} shows a question in the TREC-6 dataset that was correctly classified by the \sysname teacher as an ``Entity'' type (ENTY).
Note that the majority voting of the four weak rules that apply to this instance (Rule 8, 24, 42, and 61) leads to an incorrect prediction of ``Human'' (HUM) type. The \sysname teacher aggregates all the heuristic rule labels and the student pseudo-label with their (computed) fidelity weights for the correct prediction. 

Refer to Table~\ref{tab:trec-examples} for more illustrative examples on how \sysname aggregates various weak supervision sources with corresponding attention weights shown in parantheses. 
In Example 1 where no rules apply, the student leverages the context of the sentence (e.g., semantics of ``president'') to predict the {\tt HUM} label. 
While in Example 2, the teacher downweights the incorrect student (as well as conflicting rules) and upweights the appropriate rule to predict the correct {\tt ENTY} label. 
In example 3, \sysname predicts the correct label {\tt ENTY} relying only on the student as both rules report noisy labels. 
%

\section{Related Work}
In this section, we discuss related work on self-training and learning with noisy labels or rules. Refer to~\citet{hedderich2020survey} for a thorough survey of approaches addressing low-resource scenarios.

%

\paragraph{Self-Training.}
\label{s:related-work-self-training}
Self-training~\cite{yarowsky1995unsupervised,nigam2000analyzing, lee2013pseudo} as one of the earliest semi-supervised learning approaches~\cite{chapelle2009semi} trains a base model (student) on a small amount of labeled data; applies it to pseudo-label (task-specific) unlabeled data; uses pseudo-labels to augment the labeled data; and re-trains the student in an iterative manner.  
Self-training has recently been shown to obtain state-of-the-art performance for tasks like image classification~\cite{li2019learning,xie2020self,zoph2020rethinking}, few-shot text classification~\cite{mukherjee2020uncertainty, wang2021adaptive}, and neural machine translation~\cite{zhang2016exploiting,he2019revisiting} and has shown complementary advantages to unsupervised pre-training~\cite{zoph2020rethinking}. 
A typical issue in self-training is error propagation from noisy pseudo-labels. This is addressed in \sysname via rule attention network that computes the fidelity of pseudo-labels instead of directly using them to re-train the student. 

\paragraph{Learning with Noisy Labels.} %
\label{s:related-work-noisy-labels}
Classification under label noise from a single source has been an active research topic~\cite{frenay2013classification}. 
A major line of research focuses on correcting noisy labels by learning label corruption matrices~\cite{patrini2017making,hendrycks2018using,zheng2019meta}. 
More related to our work are the instance re-weighting approaches~\cite{ren2018learning, shu2019meta}, which learn to up-weight and down-weight instances with cleaner and noisy labels respectively. However, these operate only at instance-level and do not consider rule-specific importance.
Our approach learns both instance- and rule-specific fidelity weights and substantially outperforms ~\citet{ren2018learning} across all datasets. 

\paragraph{Learning with Multiple Rules.}
\label{s:related-work-weak-supervision}
To address the challenges with multiple noisy rules, existing approaches learn rule weights based on mutual rule agreements %
with some strong assumptions. 
For instance,~\citet{meng2018weakly,karamanolakis2019leveraging,mekala2020contextualized} denoise seed words using vector representations of their semantics. However it is difficult to generalize these approaches from seed words to more general labeling functions that only predict heuristic labels (as in our datasets). 
\citet{ratner2017snorkel, sachan2018learning, ratner2019training} assume each rule to be equally accurate across all the instances that it covers. 
~\citet{awasthi2019learning} learn rule-specific and instance-specific weights but assume access to {\em labeled exemplars} that were used to create the rule in the first place. 
Most importantly, all these works ignore unlabeled instances that are not covered by any of the rules, while our approach leverages all unlabeled instances via self-training. 

\section{Conclusions and Future Work}
We developed a weak supervision framework, \sysname, that efficiently trains classifiers by integrating task-specific unlabeled data, few labeled data, and domain-specific knowledge expressed as rules. Our framework improves data coverage by employing self-training with a student model. This considers contextualized representations of instances and predicts pseudo-labels for all instances, including those that are not covered by heuristic rules. 
Additionally, we developed a rule attention network, RAN, to aggregate various weak sources of supervision (heuristic rules and student pseudo-labels) with instance-specific weights, and employed a semi-supervised objective for training RAN without strong assumptions about the nature or structure of the weak sources. 
Extensive experiments on several benchmark datasets demonstrate our effectiveness, particularly at high levels of rule sparsity.
In future work, we plan to extend our framework to support a broader range of natural language understanding tasks and explore alternative techniques for rule embedding.

\section*{Acknowledgements}
We thank the anonymous reviewers for their constructive feedback, and Wei Wang and Benjamin Van Durme for insightful discussions. 
\newpage

\section*{Ethical Considerations}

In this work, we introduce a framework for training of neural network models with few labeled examples and domain-specific knowledge. This work is likely to increase the progress of NLP applications for domains with limited annotated resources but access to domain-specific knowledge. While it is not only expensive to acquire large amounts of labeled data for every task and language, in many cases, we cannot perform large-scale labeling due to access constraints from privacy and compliance concerns. To this end, our framework can be used for applications in finance, legal, healthcare, retail and other domains where adoption of deep neural network may have been hindered due to lack of large-scale manual annotations on sensitive data.  

While our framework accelerates the progress of NLP, it also suffers from associated societal implications of automation ranging from job losses for workers who provide annotations as a service. Additionally, it involves deep neural models that are compute intensive and has a negative impact on the environment in terms of carbon footprint. The latter concern is partly alleviated in our work by leveraging pre-trained language models and not training from scratch, thereby, leading to efficient and faster compute.

\bibliography{anthology,custom}

\begin{thebibliography}{39}
\expandafter\ifx\csname natexlab\endcsname\relax\def\natexlab#1{#1}\fi

\bibitem[{Augenstein et~al.(2016)Augenstein, Rockt{\"a}schel, Vlachos, and
  Bontcheva}]{augenstein2016stance}
Isabelle Augenstein, Tim Rockt{\"a}schel, Andreas Vlachos, and Kalina
  Bontcheva. 2016.
\newblock Stance detection with bidirectional conditional encoding.
\newblock In \emph{Proceedings of the 2016 Conference on Empirical Methods in
  Natural Language Processing}, pages 876--885.

\bibitem[{Awasthi et~al.(2020)Awasthi, Ghosh, Goyal, and
  Sarawagi}]{awasthi2019learning}
Abhijeet Awasthi, Sabyasachi Ghosh, Rasna Goyal, and Sunita Sarawagi. 2020.
\newblock Learning from rules generalizing labeled exemplars.
\newblock In \emph{International Conference on Learning Representations}.

\bibitem[{Bach et~al.(2019)Bach, Rodriguez, Liu, Luo, Shao, Xia, Sen, Ratner,
  Hancock, Alborzi et~al.}]{bach2019snorkel}
Stephen~H Bach, Daniel Rodriguez, Yintao Liu, Chong Luo, Haidong Shao,
  Cassandra Xia, Souvik Sen, Alex Ratner, Braden Hancock, Houman Alborzi,
  et~al. 2019.
\newblock Snorkel drybell: A case study in deploying weak supervision at
  industrial scale.
\newblock In \emph{Proceedings of the 2019 International Conference on
  Management of Data}.

\bibitem[{Badene et~al.(2019)Badene, Thompson, Lorr{\'e}, and
  Asher}]{badene2019data}
Sonia Badene, Kate Thompson, Jean-Pierre Lorr{\'e}, and Nicholas Asher. 2019.
\newblock Data programming for learning discourse structure.
\newblock In \emph{Association for Computational Linguistics (ACL)}.

\bibitem[{Chapelle et~al.(2009)Chapelle, Scholkopf, and
  Zien}]{chapelle2009semi}
Olivier Chapelle, Bernhard Scholkopf, and Alexander Zien. 2009.
\newblock Semi-supervised learning (chapelle, o. et al., eds.; 2006)[book
  reviews].
\newblock \emph{IEEE Transactions on Neural Networks}, 20(3):542--542.

\bibitem[{Devlin et~al.(2019)Devlin, Chang, Lee, and
  Toutanova}]{devlin2019bert}
Jacob Devlin, Ming-Wei Chang, Kenton Lee, and Kristina Toutanova. 2019.
\newblock {BERT}: Pre-training of deep bidirectional transformers for language
  understanding.
\newblock In \emph{NAACL-HLT}.

\bibitem[{Fr{\'e}nay and Verleysen(2013)}]{frenay2013classification}
Beno{\^\i}t Fr{\'e}nay and Michel Verleysen. 2013.
\newblock Classification in the presence of label noise: a survey.
\newblock \emph{IEEE transactions on neural networks and learning systems},
  25(5):845--869.

\bibitem[{Ganchev et~al.(2010)Ganchev, Gra{\c{c}}a, Gillenwater, and
  Taskar}]{ganchev2010posterior}
Kuzman Ganchev, Joao Gra{\c{c}}a, Jennifer Gillenwater, and Ben Taskar. 2010.
\newblock Posterior regularization for structured latent variable models.
\newblock \emph{The Journal of Machine Learning Research}, 11:2001--2049.

\bibitem[{Grandvalet and Bengio(2005)}]{grandvalet2005semi}
Yves Grandvalet and Yoshua Bengio. 2005.
\newblock Semi-supervised learning by entropy minimization.
\newblock In \emph{Advances in Neural Information Processing Systems}.

\bibitem[{He et~al.(2019)He, Gu, Shen, and Ranzato}]{he2019revisiting}
Junxian He, Jiatao Gu, Jiajun Shen, and Marc'Aurelio Ranzato. 2019.
\newblock Revisiting self-training for neural sequence generation.
\newblock In \emph{International Conference on Learning Representations}.

\bibitem[{Hedderich et~al.(2021)Hedderich, Lange, Adel, Str{\"o}tgen, and
  Klakow}]{hedderich2020survey}
Michael~A Hedderich, Lukas Lange, Heike Adel, Jannik Str{\"o}tgen, and Dietrich
  Klakow. 2021.
\newblock A survey on recent approaches for natural language processing in
  low-resource scenarios.
\newblock In \emph{Proceedings of NAACL-HLT}.

\bibitem[{Hendrycks et~al.(2018)Hendrycks, Mazeika, Wilson, and
  Gimpel}]{hendrycks2018using}
Dan Hendrycks, Mantas Mazeika, Duncan Wilson, and Kevin Gimpel. 2018.
\newblock Using trusted data to train deep networks on labels corrupted by
  severe noise.
\newblock In \emph{Advances in Neural Information Processing Systems}, pages
  10477--10486.

\bibitem[{Hu et~al.(2016)Hu, Ma, Liu, Hovy, and Xing}]{hu2016harnessing}
Zhiting Hu, Xuezhe Ma, Zhengzhong Liu, Eduard Hovy, and Eric Xing. 2016.
\newblock Harnessing deep neural networks with logic rules.
\newblock In \emph{Proceedings of the 54th Annual Meeting of the Association
  for Computational Linguistics}.

\bibitem[{Karamanolakis et~al.(2019)Karamanolakis, Hsu, and
  Gravano}]{karamanolakis2019leveraging}
Giannis Karamanolakis, Daniel Hsu, and Luis Gravano. 2019.
\newblock Leveraging just a few keywords for fine-grained aspect detection
  through weakly supervised co-training.
\newblock In \emph{Proceedings of the 2019 Conference on Empirical Methods in
  Natural Language Processing and the 9th International Joint Conference on
  Natural Language Processing (EMNLP-IJCNLP)}.

\bibitem[{Lee(2013)}]{lee2013pseudo}
Dong-Hyun Lee. 2013.
\newblock Pseudo-label: The simple and efficient semi-supervised learning
  method for deep neural networks.
\newblock In \emph{Workshop on challenges in representation learning, ICML}.

\bibitem[{Li et~al.(2019)Li, Sun, Liu, Zhou, Zheng, Chua, and
  Schiele}]{li2019learning}
Xinzhe Li, Qianru Sun, Yaoyao Liu, Qin Zhou, Shibao Zheng, Tat-Seng Chua, and
  Bernt Schiele. 2019.
\newblock Learning to self-train for semi-supervised few-shot classification.
\newblock In \emph{Advances in Neural Information Processing Systems}.

\bibitem[{Mekala and Shang(2020)}]{mekala2020contextualized}
Dheeraj Mekala and Jingbo Shang. 2020.
\newblock Contextualized weak supervision for text classification.
\newblock In \emph{Proceedings of the 58th Annual Meeting of the Association
  for Computational Linguistics}.

\bibitem[{Meng et~al.(2018)Meng, Shen, Zhang, and Han}]{meng2018weakly}
Yu~Meng, Jiaming Shen, Chao Zhang, and Jiawei Han. 2018.
\newblock Weakly-supervised neural text classification.
\newblock In \emph{Proceedings of the 27th ACM International Conference on
  Information and Knowledge Management}.

\bibitem[{Mintz et~al.(2009)Mintz, Bills, Snow, and
  Jurafsky}]{mintz2009distant}
Mike Mintz, Steven Bills, Rion Snow, and Dan Jurafsky. 2009.
\newblock Distant supervision for relation extraction without labeled data.
\newblock In \emph{Proceedings of the Joint Conference of the 47th Annual
  Meeting of the ACL and the 4th International Joint Conference on Natural
  Language Processing of the AFNLP}.

\bibitem[{Mukherjee and Awadallah(2020)}]{mukherjee2020uncertainty}
Subhabrata Mukherjee and Ahmed Awadallah. 2020.
\newblock Uncertainty-aware self-training for few-shot text classification.
\newblock In \emph{Advances in Neural Information Processing Systems}.

\bibitem[{Nigam and Ghani(2000)}]{nigam2000analyzing}
Kamal Nigam and Rayid Ghani. 2000.
\newblock Analyzing the effectiveness and applicability of co-training.
\newblock In \emph{Proceedings of the ninth international conference on
  Information and knowledge management}.

\bibitem[{Patrini et~al.(2017)Patrini, Rozza, Krishna~Menon, Nock, and
  Qu}]{patrini2017making}
Giorgio Patrini, Alessandro Rozza, Aditya Krishna~Menon, Richard Nock, and
  Lizhen Qu. 2017.
\newblock Making deep neural networks robust to label noise: A loss correction
  approach.
\newblock In \emph{Proceedings of the IEEE Conference on Computer Vision and
  Pattern Recognition}, pages 1944--1952.

\bibitem[{Peters et~al.(2018)Peters, Neumann, Iyyer, Gardner, Clark, Lee, and
  Zettlemoyer}]{peters2018deep}
Matthew~E Peters, Mark Neumann, Mohit Iyyer, Matt Gardner, Christopher Clark,
  Kenton Lee, and Luke Zettlemoyer. 2018.
\newblock Deep contextualized word representations.
\newblock In \emph{Proceedings of NAACL-HLT}.

\bibitem[{Platanios et~al.(2017)Platanios, Poon, Mitchell, and
  Horvitz}]{platanios2017estimating}
Emmanouil Platanios, Hoifung Poon, Tom~M Mitchell, and Eric~J Horvitz. 2017.
\newblock Estimating accuracy from unlabeled data: A probabilistic logic
  approach.
\newblock \emph{Advances in Neural Information Processing Systems}.

\bibitem[{Radford et~al.(2019)Radford, Wu, Child, Luan, Amodei, and
  Sutskever}]{radford2019language}
Alec Radford, Jeffrey Wu, Rewon Child, David Luan, Dario Amodei, and Ilya
  Sutskever. 2019.
\newblock Language models are unsupervised multitask learners.
\newblock \emph{OpenAI blog}, 1(8):9.

\bibitem[{Ratner et~al.(2017)Ratner, Bach, Ehrenberg, Fries, Wu, and
  R{\'e}}]{ratner2017snorkel}
Alexander Ratner, Stephen~H Bach, Henry Ehrenberg, Jason Fries, Sen Wu, and
  Christopher R{\'e}. 2017.
\newblock Snorkel: Rapid training data creation with weak supervision.
\newblock In \emph{Proceedings of the VLDB Endowment. International Conference
  on Very Large Data Bases}, volume~11, page 269. NIH Public Access.

\bibitem[{Ratner et~al.(2019)Ratner, Hancock, Dunnmon, Sala, Pandey, and
  R{\'e}}]{ratner2019training}
Alexander Ratner, Braden Hancock, Jared Dunnmon, Frederic Sala, Shreyash
  Pandey, and Christopher R{\'e}. 2019.
\newblock Training complex models with multi-task weak supervision.
\newblock In \emph{Proceedings of the AAAI Conference on Artificial
  Intelligence}.

\bibitem[{Ren et~al.(2018{\natexlab{a}})Ren, Stewart, Song, Kuleshov, and
  Ermon}]{ren2018learningb}
Hongyu Ren, Russell Stewart, Jiaming Song, Volodymyr Kuleshov, and Stefano
  Ermon. 2018{\natexlab{a}}.
\newblock Learning with weak supervision from physics and data-driven
  constraints.
\newblock \emph{AI Magazine}, 39(1):27--38.

\bibitem[{Ren et~al.(2018{\natexlab{b}})Ren, Zeng, Yang, and
  Urtasun}]{ren2018learning}
Mengye Ren, Wenyuan Zeng, Bin Yang, and Raquel Urtasun. 2018{\natexlab{b}}.
\newblock Learning to reweight examples for robust deep learning.
\newblock In \emph{International Conference on Machine Learning}.

\bibitem[{Riloff(1996)}]{riloff1996automatically}
Ellen Riloff. 1996.
\newblock Automatically generating extraction patterns from untagged text.
\newblock In \emph{Proceedings of the national conference on artificial
  intelligence}.

\bibitem[{Sachan et~al.(2018)Sachan, Dubey, Mitchell, Roth, and
  Xing}]{sachan2018learning}
Mrinmaya Sachan, Kumar~Avinava Dubey, Tom~M Mitchell, Dan Roth, and Eric~P
  Xing. 2018.
\newblock Learning pipelines with limited data and domain knowledge: A study in
  parsing physics problems.
\newblock In \emph{Advances in Neural Information Processing Systems}.

\bibitem[{Shu et~al.(2019)Shu, Xie, Yi, Zhao, Zhou, Xu, and Meng}]{shu2019meta}
Jun Shu, Qi~Xie, Lixuan Yi, Qian Zhao, Sanping Zhou, Zongben Xu, and Deyu Meng.
  2019.
\newblock Meta-weight-net: Learning an explicit mapping for sample weighting.
\newblock In \emph{Advances in Neural Information Processing Systems}.

\bibitem[{Wang et~al.(2020)Wang, Mukherjee, Chu, Tu, Wu, Gao, and
  Awadallah}]{wang2021adaptive}
Yaqing Wang, Subhabrata Mukherjee, Haoda Chu, Yuancheng Tu, Ming Wu, Jing Gao,
  and Ahmed Awadallah. 2020.
\newblock Adaptive self-training for few-shot neural sequence labeling.
\newblock \emph{arXiv preprint arXiv:2010.03680}.

\bibitem[{Xie et~al.(2020)Xie, Luong, Hovy, and Le}]{xie2020self}
Qizhe Xie, Minh-Thang Luong, Eduard Hovy, and Quoc~V Le. 2020.
\newblock Self-training with noisy student improves imagenet classification.
\newblock In \emph{Proceedings of the IEEE/CVF Conference on Computer Vision
  and Pattern Recognition}, pages 10687--10698.

\bibitem[{Xu et~al.(2013)Xu, Hoffmann, Zhao, and Grishman}]{xu2013filling}
Wei Xu, Raphael Hoffmann, Le~Zhao, and Ralph Grishman. 2013.
\newblock Filling knowledge base gaps for distant supervision of relation
  extraction.
\newblock In \emph{Proceedings of the 51st Annual Meeting of the Association
  for Computational Linguistics}.

\bibitem[{Yarowsky(1995)}]{yarowsky1995unsupervised}
David Yarowsky. 1995.
\newblock Unsupervised word sense disambiguation rivaling supervised methods.
\newblock In \emph{33rd annual meeting of the association for computational
  linguistics}, pages 189--196.

\bibitem[{Zhang and Zong(2016)}]{zhang2016exploiting}
Jiajun Zhang and Chengqing Zong. 2016.
\newblock Exploiting source-side monolingual data in neural machine
  translation.
\newblock In \emph{Proceedings of the 2016 Conference on Empirical Methods in
  Natural Language Processing}.

\bibitem[{Zheng et~al.(2021)Zheng, Awadallah, and Dumais}]{zheng2019meta}
Guoqing Zheng, Ahmed~Hassan Awadallah, and Susan Dumais. 2021.
\newblock Meta label correction for noisy label learning.
\newblock In \emph{Proceedings of the 35th AAAI Conference on Artificial
  Intelligence}.

\bibitem[{Zoph et~al.(2020)Zoph, Ghiasi, Lin, Cui, Liu, Cubuk, and
  Le}]{zoph2020rethinking}
Barret Zoph, Golnaz Ghiasi, Tsung-Yi Lin, Yin Cui, Hanxiao Liu, Ekin~Dogus
  Cubuk, and Quoc Le. 2020.
\newblock Rethinking pre-training and self-training.
\newblock \emph{Advances in Neural Information Processing Systems}.

\end{thebibliography}
\bibliographystyle{acl_natbib}

\newpage 
\appendix
\section{Appendix}
\label{sec:appendix}
For reproducibility, we provide details of our implementation (Section~\ref{appendix-implementation-details}), datasets (Section~\ref{appendix-dataset-details}), and experimental results (Section~\ref{appendix-experiment-details}).
Our code is available at~\url{https://github.com/microsoft/ASTRA}.

\subsection{Implementation Details}
\label{appendix-implementation-details}
We now describe implementation details for each component in \sysname: our base student model and our rule attention teacher network. 
Table~\ref{tab:hyperparameters} shows our hyperparameter search configuration. We choose optimal hyperparameters by manual tuning based on the development performances.
Table~\ref{tab:best-hyperparameters} shows the hyperparameters and model architecture details for each dataset. 
For a fair comparison, we use the same architectures as previous approaches but we expect further improvements by exploring different architectures.

\paragraph{Base Student Model}
Our student model consists of an instance embedding layer (e.g., ELMO~\cite{peters2018deep}, BERT~\cite{devlin2019bert}, logistic regression), a multilayer perceptron with two hidden layers, and a softmax classification layer for predicting labels.

\paragraph{Rule Attention Teacher Network}
Our RAN teacher model consists of a 128-dimensional rule embedding layer, a multilayer perceptron for mapping the contextualized representation for an instance to the rule embedding space, and a sigmoid attention layer.  

\paragraph{Iterative Teacher-Student Training}
At each iteration, we train the RAN teacher on unlabeled data and fine-tune on clean labeled data.
Also at each iteration, we train the student on pseudo-labeled teacher data and fine-tune on clean labeled data.
We consider a maximum number of 25 self-training iterations (with early stopping of patience 3 epochs) and keep the models' performances for the iteration corresponding to the highest validation performance. 
 
\subsection{Dataset Details}
\label{appendix-dataset-details}
 We evaluate our framework on the following six benchmark datasets for weak supervision from ~\citet{ratner2017snorkel} and~\citet{awasthi2019learning}\footnote{\url{https://github.com/awasthiabhijeet/Learning-From-Rules}}.
 All datasets are in English. 
 Table~\ref{tab:dataset-statistics-detailed} shows detailed dataset statistics.
 We consider the same test sets with previous work. 
 For a robust evaluation of our model's performance, we split each dataset into five random train/validation/unlabeled splits and report the average performance and standard deviation across runs. For a fair comparison, we use the same splits and evaluation procedure across all methods and baselines. 
 
 \paragraph{TREC:} Question classification from TREC-6 into $6$ categories: Abbreviation, Entity, Description, Human, Location, Numeric-value. Table~\ref{tab:trec_rules_sample} reports a sample of regular expression rules out of the 68 rules used in the TREC dataset.  TREC has 13 keyword-based (coverage=62\%) and 55 regular expression-based (coverage=57\%) rules. 
 
 \paragraph{SMS:} Binary Spam vs. Not Spam classification of SMS messages. SMS has 16 keyword-based (coverage=4\%) and 57 regular expression-based (coverage=38\%) rules.
 \paragraph{YouTube:} Binary Spam vs. Not Spam classification of YouTube comments.\footnote{\url{https://archive.ics.uci.edu/ml/machine-learning-databases/00380/YouTube-Spam-Collection-v1.zip}}
 YouTube has 5 keyword-based (coverage=48\%), 1 regular expression-based (coverage=23\%), 1 length-based (coverage=23\%), and 3 classifier-based (coverage=46\%) rules. 
 \paragraph{CENSUS:} Binary income classification on the UCI CENSUS dataset on whether a person earns more than \$50K or not. This is a non-textual dataset and is considered to evaluate the performance of our approach under the low sparsity setting, since the 83 rules are automatically extracted and have a coverage of 100\%.
 \paragraph{MIT-R:} Slot-filling in sentences on restaurant search queries in the MIT-R dataset: each token is classified into $9$ classes (Location, Hours, Amenity, Price, Cuisine, Dish, Restaurant Name, Rating, Other). MIT-R has 5 keyword-based (coverage=6\%) and 10 regular expression-based (coverage=10\%) rules. 
 \paragraph{Spouse:} Relation classification in the Spouses dataset\footnote{\url{https://www.dropbox.com/s/jmrvyaqew4zp9cy/spouse_data.zip}}, whether pairs of people mentioned in a sentence are/were married or not. Spouse has 6 keyword-based (coverage=23\%), 1 heuristic-based (coverage=4\%), and 2 distant supervision-based (coverage=0.2\%) rules.

\subsection{Experimental Result Details}
\label{appendix-experiment-details}
We now discuss detailed results on each dataset. 
To be consistent with previous work, we report accuracy scores for the TREC, Youtube, and CENSUS dataset and macro-average F1 scores for the SMS, Spouse, and MIT-R datasets. 

\subsubsection{Ablation Studies}
Table~\ref{tab:ablation-results-detailed} reports detailed ablation results per dataset. The right column computes the average accuracy across datasets. 

\subsubsection{Case Study: TREC-6 Dataset}
Table~\ref{tab:trec_rules_sample} shows a sample of rules from the TREC-6 dataset. Those rules capture regular expression patterns to predict one of the 6 question categories for a question. 
Tables~\ref{tab:trec_example_first}-\ref{tab:trec_example_last} show examples of individual instances in TREC-6, the corresponding rule predictions, the student pseudo-labels, as well as our RAN  that aggregates rule and student predictions with attention weights $a$ to compute a single (Teacher) label. 

\begin{table*}[]
    \centering
    \begin{tabular}{ll}
        \toprule 
        \textbf{Hyperparameter} & \textbf{Values} \\\midrule  
         Learning rate & 1e-1, 1e-2, 1e-3, 1e-4, 1e-5\\
         Fine-tuning rate &  1e-3, 1e-4, 1e-5, 1e-6, 1e-7\\
         Type of pseudo-labels &  soft, hard\\
         \bottomrule 
    \end{tabular}
    \caption{Hyperparameter search.}
    \label{tab:hyperparameters}
\end{table*}

\begin{table*}[]
\centering
\resizebox{2\columnwidth}{!}{
\begin{tabular}{lcccccc}
\toprule  
                       & \textbf{TREC}  & \textbf{SMS}   & \textbf{Youtube} & \textbf{CENSUS}  & \textbf{MIT-R} &  \textbf{Spouse}   \\\midrule 

Instance vector type & ELMO & ELMO  & LogReg &  Categorical & ELMO & BERT\\
Instance vector dimensionality & 1024 &1024  & 16,634 &  105 & 1024 & 768\\
Learning rate & 1e-4 &1e-4  &  1e-4 &  1e-3 &   1e-4 & 1e-4\\
Fine-tuning rate & 1e-5 &1e-5  &  1e-5&  1e-5 &   1e-4 & 1e-5\\
Type of pseudo-labels & soft &soft &  hard & soft &   soft & soft\\
Pseudo-training epochs (patience: 5) & 25 &25 &  25 & 25 &   25 & 25\\
Fine-tuning epochs (patience: 5) & 70 &70 &  70 & 70 &   70 & 70\\
Self-training iterations (patience: 3)& 25 &25 &  25 & 25 &   25 & 25\\
Training batch size & 16 &16 &  16 & 128 &   256 & 16\\
Unsupervised batch size & 256 &256 &  256 & 256 &   256 & 256\\
Rule embedding dimensionality & 128 &128 &  128 & 128 &   128 & 128\\
\bottomrule 
\end{tabular}}
\caption{Selected hyperparameters.}
\label{tab:best-hyperparameters}
\end{table*}

\begin{table*}[]
\centering
\resizebox{2\columnwidth}{!}{
\begin{tabular}{lcccccc}
\toprule  
                       & \textbf{TREC}  & \textbf{SMS}   & \textbf{Youtube} & \textbf{CENSUS}  & \textbf{MIT-R}  & \textbf{AVG} \\
                      & \textbf{ (Acc)} & \textbf{(F1)}   & \textbf{(Acc)} & \textbf{(Acc)} &  \textbf{(F1)}  & \textbf{(Acc)} \\\midrule  
\sysname (Teacher) & \textbf{80.3} (2.4) & 95.3 (0.5) & 95.3 (0.8)    & \textbf{83.1} (0.4)   & \textbf{76.9} (0.6) &  \textbf{88.1}  \\
\sysname (Student) & 79.2 (2.1) & \textbf{95.7} (0.5)& \textbf{95.5} (0.5)    & 82.8 (0.1)   & 76.6 (0.9) &  87.7   \\
Hard Student Pseudo-labels in RAN in Eq. (2) & 77.6 (1.2)	& 94.5 (0.7) & 95.3 (0.8) &	83.0 (0.7)	& 75.9 (0.9) &  87.0\\
No minimum entropy regularization in Eq. (4) & 75.8 (2.5)&	\textbf{95.7} (0.7)	& 95.1 (0.8) &	83.0 (0.3) &	72.6 (0.4) & 86.9\\
No cross-entropy loss in Eq. (4) & 74.1 (2.3) & 93.9 (0.4)	& \textbf{95.5} (0.6)	& \textbf{83.1} (0.6)	& 71.9 (0.4)& 86.7\\
No student pseudo-labels in RAN in Eq. (1) & 75.3 (4.3)	& 95.8 (0.2)	& 91.4 (2.2) &	\textbf{83.1} (0.4)	& 71.9 (1.0) & 85.3 \\

\bottomrule 
\end{tabular}}
\caption{Ablation studies.}
\label{tab:ablation-results-detailed}
\end{table*}

\begin{table*}[!htb]
\resizebox{2\columnwidth}{!}{
\begin{tabular}{llllllll}
\toprule 
& \textbf{TREC}  & \textbf{SMS} &\textbf{Youtube} &  \textbf{CENSUS} & \textbf{MIT-R}  & \textbf{Spouse}  \\\midrule 
$|D_L|$ & 68      & 69 & 100      & 83   & 1842     &  100 \\
$|D_U|$ & 4884    & 1586     & 4502    & 64,888    & 10,000    & 22,254 \\
Validation & 500  & 500  & 150   & 5561   & 16281      & 2711 \\
Test Size & 500  & 500  & 250   & 16281   & 14256      & 2701 \\
\#Classes & 6    & 2     & 2      & 2 & 9  & 2 \\
\#Rules & 68       & 73   & 10   & 83   & 15     & 9 \\
Rule Precision (Majority Voting) & 63.7\% & 97.3\%& 78.6\% & 80.7\%& 84.1\% &  66.6\%\\
Rule Accuracy (Majority Voting) & 60.9\% & 48.4\%& 82.2\% & 80.1\%& 40.9\% &  44.2\%\\
Rule Coverage (instances in $D_U$ covered by $\geq 1$ rule)& 95\% & 40\%& 87\%  & 100\% & 14\%   & 25\%\\
Rule Overlap (instances in $D_U$ covered by $\geq 2$ rules)& 46\%  & 9\% & 48\%  & 94\%& 1\%   & 8\%\\ 
\bottomrule   
\end{tabular}}
\caption{Dataset statistics. }
\label{tab:dataset-statistics-detailed}
\end{table*}

\begin{table*}[]
    \centering
    \begin{tabular}{lll}
    \toprule
    \textbf{Rule} & \textbf{Label} & \textbf{Pattern}\\\midrule
            Rule 5 & HUM & {\small \verb!( |^)(which|who|what|what)[^\w]*([^\s]+ )*(person|!} \\
         &   & {\small \verb! man|woman|human|poet|poet)[^\w]*( |$)!} \\\hline 
    
    Rule 8 & HUM  & {\small \verb!( |^)(who|what|what)[^\w] *(\w+ ){0,1}(person|!} \\
         &   & {\small \verb!man|woman|human|president|president)[^\w]*( |$)!} \\\hline 
    Rule 24 & ENTY   & {\small \verb!( |^)(what|what)[^\w]*(\w+ ){0,1}(is|is)[^\w]*!}\\
      & & {\small \verb!*([^\s]+ )*(surname|address|name|name)[^\w]*( |$)!}  \\\hline 
     Rule 29 & NUM  &  {\small \verb!( |^)(which|what|what)[^\w]* !} \\
           &   & {\small \verb!*([^\s]+ )*(time|day|month|hours|minute!} \\
      &   & {\small \verb!*|seconds|year|date|date)[^\w]*( |$)!} \\\hline 
         Rule 32 & NUM  & {\small \verb!( |^)(year|year)[^\w]*( |$)!} \\\hline
    Rule 41 &NUM & {\small \verb!( |^)(what|what)[^\w]* ([^\s]+ )*(percentage!} \\
         &   & {\small \verb!|share|number|population|population)[^\w]*( |$)!} \\\hline 
    Rule 42 & DESC  & {\small \verb!( |^)(explain|describe|what|what)[^\w]*( |$)!} \\\hline 
     Rule 54 &DESC  & {\small \verb!( |^)(how|what|what)[^\w]*!} \\
      & &  {\small \verb!* (\w+ ){0,1}(do|does|does)[^\w]*( |$)!} \\\hline 
              Rule 61 & HUM & {\small \verb!( |^)(called|alias|nicknamed|nicknamed)[^\w]*( |$)!} \\\hline
        Rule 68 & ABBR  & {\small \verb!( |^)(what|what)[^\w]* (\w+ ){0,1}(does|does)[^\w]*!} \\ 
      &  & {\small \verb!* * ([^\s]+ )*(stand for)[^\w]*( |$)!} \\
       \bottomrule 
    \end{tabular}
    \caption{Sample of REGEX rules from the TREC-6 dataset capturing the various question categories (HUM: Human, ENTY: Entity, NUM: Numeric Value, DESC: Description, ABBR: Abbreviation)}
    \label{tab:trec_rules_sample}
\end{table*}

\begin{table*}[!htb]
    \centering
    \begin{tabular}{llll}
    \toprule  
        Text & \multicolumn{3}{l}{Why is a ladybug helpful ?}  \\
        Clean label & DESC &&\\
        RAN Teacher & \color{forestgreen}{\textbf{DESC}} &&\\\midrule
        \textbf{Weak Source} & \textbf{Label} & \textbf{Weight}  & \textbf{Feature}\\

        Student & \color{forestgreen}{\textbf{DESC}} &  $a$=$1.0$ & $h_i$ (contextualized instance embedding)\\
    \bottomrule 
    \end{tabular}
    \caption{TREC example. No rules apply. The student generalizes beyond rules by considering contextualized instance embeddings and assigns the instance to the DESCRIPTION class. Our RAN teacher assigns an attention weight of $1$ to the student and predicts the right class.}
    \label{tab:trec_example_first}
\end{table*}

\begin{table*}[!htb]
    \centering
    \begin{tabular}{llll}
    \toprule  
        Text & \multicolumn{3}{l}{Which president was unmarried ?}  \\
        Clean label & HUM &&\\
        RAN Teacher & \color{forestgreen}{\textbf{NUM}} &&\\\midrule
        \textbf{Weak Source} & \textbf{Label} & \textbf{Weight}  & \textbf{Feature}\\

        Student & \color{forestgreen}{\textbf{HUM}} &  $a$=$1.0$ & $h_i$ (contextualized instance embedding)\\    \bottomrule 
    \end{tabular}
    \caption{TREC example: While no rules apply to this instance, the student leverages the semantics of the sentence (i.e., that ``president'' corresponds to a person) to predict the HUMAN (HUM) class. Thus, our RAN predicts the right label instead of discarding this instance.}
    \label{tab:my_label}
\end{table*}

\begin{table*}[!htb]
    \centering
    \begin{tabular}{llll}
    \toprule  
        Text & \multicolumn{3}{l}{What is a baby turkey called ?}  \\
        Clean label & ENTY &&\\
        RAN Teacher & \color{forestgreen}{\textbf{ENTY}} &&\\\midrule
        \textbf{Weak Source} & \textbf{Label} & \textbf{Weight}  & \textbf{Feature}\\

        Student & \color{red}{\textbf{DESC}} &  $a$=$1.0$ & $h_i$ (contextualized instance embedding)\\
        Rule 24 & \color{forestgreen}{\textbf{ENTY}} & $a$=$1.0$ & {\small \verb!( |^)(what|what)[^\w]*(\w+ ){0,1}(is|is)[^\w]*!} \\
         &  &  & {\small \verb!*([^\s]+ )*(surname|address|name|name)[^\w]*( |$)!} \\
         Rule 42 & \color{red}{\textbf{DESC}}  & $a$=$0.0$ & {\small \verb!( |^)(explain|describe|what|what)[^\w]*( |$)!} \\
         Rule 61 & \color{red}{\textbf{HUM}} & $a$=$0.0$ & {\small \verb!( |^)(called|alias|nicknamed|nicknamed)[^\w]*( |$)!} \\
    \bottomrule 
    \end{tabular}
    \caption{TREC example: Student is wrong but unsure. Teacher predicts the right label and fixes student's mistake. Teacher correctly down-weights rule 42 and rule 61 that provide the wrong prediction but erroneously up-weights the Student. As in this case the Student is uncertain, about the label, the final aggregated prediction of the Teacher is mostly influenced by Rule 24.}
    \label{tab:my_label}
\end{table*}

\begin{table*}[!htb]
    \centering
    \begin{tabular}{llll}
    \toprule  
        Text & \multicolumn{3}{l}{What currency do they use in Brazil ?}  \\
        Clean label & ENTY &&\\
        RAN Teacher & \color{forestgreen}{\textbf{ENTY}} &&\\\midrule
        \textbf{Weak Source} & \textbf{Label} & \textbf{Weight}  & \textbf{Feature}\\

        Student & \color{forestgreen}{\textbf{ENTY}} &  $a$=$1.0$ & $h_i$ (contextualized instance embedding)\\
         Rule 42 & \color{red}{\textbf{DESC}}  & $a$=$0.0$ & {\small \verb!( |^)(explain|describe|what|what)[^\w]*( |$)!} \\
         Rule 54 & \color{red}{\textbf{DESC}} & $a$=$0.0$ & {\small \verb!( |^)(how|what|what)[^\w]*!} \\
      &  &  & {\small \verb!* (\w+ ){0,1}(do|does|does)[^\w]*( |$)!} \\
    \bottomrule 
    \end{tabular}
    \caption{TREC example. The student is crucial for RAN to predict the right label (ENTITY) as both rules predict the wrong label. RAN correctly down-weights the two rules and up-weights the Student.}
    \label{tab:my_label}
\end{table*}

\begin{table*}[!htb]
    \centering
    \begin{tabular}{llll}
    \toprule  
        Text & \multicolumn{3}{l}{What is an atom ?}  \\
        Clean label & DESC &&\\
        RAN Teacher & \color{forestgreen}{\textbf{DESC}} &&\\\midrule
        \textbf{Weak Source} & \textbf{Label} & \textbf{Weight}  & \textbf{Feature}\\

        Student & \color{forestgreen}{\textbf{DESC}} &  $a$=$1.0$ & $h_i$ (contextualized instance embedding)\\
         Rule 42 & \color{forestgreen}{\textbf{DESC}}  & $a$=$1.0$ & {\small \verb!( |^)(explain|describe|what|what)[^\w]*( |$)!} \\
    \bottomrule 
    \end{tabular}
    \caption{TREC example: Rule 42 was down-weighted in the previous two examples but is up-weighted here, demonstrating that RAN effectively leverages the contextualized instance representation to predict instance-specific rule weights.}
    \label{tab:my_label}
\end{table*}

\begin{table*}[!htb]
    \centering
    \begin{tabular}{llll}
    \toprule  
        Text & \multicolumn{3}{l}{What was President Lyndon Johnson 's reform program called ?}  \\
        Clean label & ENTY &&\\
        RAN Teacher & \color{forestgreen}{\textbf{ENTY}} &&\\\midrule
        \textbf{Weak Source} & \textbf{Label} & \textbf{Weight}  & \textbf{Feature}\\
        Student & \color{forestgreen}{\textbf{ENTY}} &  $a$=$1.0$ & $h_i$ (contextualized instance embedding)\\
        Rule 8 & \color{red}{\textbf{HUM}} & $a$=$1.0$ & {\small \verb!( |^)(who|what|what)[^\w] *(\w+ ){0,1}(person|!} \\
         &  &  & {\small \verb!man|woman|human|president|president)[^\w]*( |$)!} \\
        Rule 24 & \color{forestgreen}{\textbf{ENTY}} & $a$=$1.0$ & {\small \verb!( |^)(what|what)[^\w]*(\w+ ){0,1}(is|is)[^\w]*!} \\
         &  &  & {\small \verb!*([^\s]+ )*(surname|address|name|name)[^\w]*( |$)!} \\
         Rule 42 & \color{red}{\textbf{DESC}}  & $a$=$0.0$ & {\small \verb!( |^)(explain|describe|what|what)[^\w]*( |$)!} \\
         Rule 61 & \color{red}{\textbf{HUM}} & $a$=$0.0$ & {\small \verb!( |^)(called|alias|nicknamed|nicknamed)[^\w]*( |$)!} \\
    \bottomrule 
    \end{tabular}
    \caption{TREC example.}
    \label{tab:my_label}
\end{table*}

\begin{table*}[!htb]
    \centering
    \begin{tabular}{llll}
    \toprule  
        Text & \multicolumn{3}{l}{What is the percentage of water content in the human body ?}  \\
        Clean label & NUM &&\\
        RAN Teacher & \color{forestgreen}{\textbf{NUM}} &&\\\midrule
        \textbf{Weak Source} & \textbf{Label} & \textbf{Weight}  & \textbf{Feature}\\
        Student & \color{forestgreen}{\textbf{DESC}} &  $a$=$0.0$ & $h_i$ (contextualized instance embedding)\\
        Rule 5 & \color{red}{\textbf{HUM}} & $a$=$0.0$ & {\small \verb!( |^)(which|who|what|what)[^\w]*([^\s]+ )*(person|!} \\
         &  &  & {\small \verb! man|woman|human|poet|poet)[^\w]*( |$)!} \\
         Rule 41 & \color{forestgreen}{\textbf{NUM}} & $a$=$0.2$ & {\small \verb!( |^)(what|what)[^\w]* ([^\s]+ )*(percentage!} \\
         &  &  & {\small \verb!|share|number|population|population)[^\w]*( |$)!} \\
         Rule 42 & \color{red}{\textbf{DESC}}  & $a$=$0.0$ & {\small \verb!( |^)(explain|describe|what|what)[^\w]*( |$)!} \\
    \bottomrule 
    \end{tabular}
    \caption{TREC example.}
    \label{tab:my_label}
\end{table*}

\begin{table*}[!htb]
    \centering
    \begin{tabular}{llll}
    \toprule  
        Text & \multicolumn{3}{l}{How fast is alcohol absorbed ?}  \\
        Clean label & NUM &&\\
        RAN Teacher & \color{forestgreen}{\textbf{NUM}} &&\\\midrule
        \textbf{Weak Source} & \textbf{Label} & \textbf{Weight}  & \textbf{Feature}\\

        Student & \color{forestgreen}{\textbf{NUM}} &  $a$=$1.0$ & $h_i$ (contextualized instance embedding)\\    \bottomrule 
    \end{tabular}
    \caption{TREC example: While no rules apply to this instance, the student associates ``How fast''  with the NUMBER (NUM) class. }
    \label{tab:my_label}
\end{table*}

\begin{table*}[!htb]
    \centering
    \begin{tabular}{llll}
    \toprule  
        Text & \multicolumn{3}{l}{Which mountain range in North America stretches from Maine to Georgia ?}  \\
        Clean label & LOC &&\\
        RAN Teacher & \color{forestgreen}{\textbf{LOC}} &&\\\midrule
        \textbf{Weak Source} & \textbf{Label} & \textbf{Weight}  & \textbf{Feature}\\

        Student & \color{forestgreen}{\textbf{LOC}} &  $a$=$1.0$ & $h_i$ (contextualized instance embedding)\\    \bottomrule 
    \end{tabular}
    \caption{TREC example: While no rules apply to this instance, the student associates the context with the LOCATION (LOC) class. }
    \label{tab:my_label}
\end{table*}

\begin{table*}[!htb]
    \centering
    \begin{tabular}{llll}
    \toprule  
        Text & \multicolumn{3}{l}{When is the official first day of summer ?}  \\
        Clean label & NUM &&\\
        RAN Teacher & \color{forestgreen}{\textbf{NUM}} &&\\\midrule
        \textbf{Weak Source} & \textbf{Label} & \textbf{Weight}  & \textbf{Feature}\\

        Student & \color{forestgreen}{\textbf{NUM}} &  $a$=$1.0$ & $h_i$ (contextualized instance embedding)\\    \bottomrule 
    \end{tabular}
    \caption{TREC example: While no rules apply to this instance, the student associates ``when,'' ``day,'' and ``summer'' to NUMBER (NUM) class. }
    \label{tab:my_label}
\end{table*}

\begin{table*}[!htb]
    \centering
    \begin{tabular}{llll}
    \toprule  
        Text & \multicolumn{3}{l}{What is Australia 's national flower ?}  \\
        Clean label & ENTY &&\\
        RAN Teacher & \color{red}{\textbf{DESC}} &&\\\midrule
        \textbf{Weak Source} & \textbf{Label} & \textbf{Weight}  & \textbf{Feature}\\
        Student & \color{red}{\textbf{DESC}} &  $a$=$1.0$ & $h_i$ (contextualized instance embedding)\\
         Rule 42 & \color{red}{\textbf{DESC}}  & $a$=$0.0$ & {\small \verb!( |^)(explain|describe|what|what)[^\w]*( |$)!} \\
    \bottomrule 
    \end{tabular}
    \caption{TREC example: Both the Student and Rule 42 provide a wrong prediction}
    \label{tab:my_label}
\end{table*}

\begin{table*}[!htb]
    \centering
    \begin{tabular}{llll}
    \toprule  
        Text & \multicolumn{3}{l}{What is the name of the chocolate company in San Francisco ?}  \\
        Clean label & HUM &&\\
        RAN Teacher & \color{forestgreen}{\textbf{ENTY}} &&\\\midrule
        \textbf{Weak Source} & \textbf{Label} & \textbf{Weight}  & \textbf{Feature}\\

        Student & \color{forestgreen}{\textbf{ENTY}} &  $a$=$1.0$ & $h_i$ (contextualized instance embedding)\\
         Rule 42 & \color{red}{\textbf{DESC}}  & $a$=$0.0$ & {\small \verb!( |^)(explain|describe|what|what)[^\w]*( |$)!} \\
          Rule 53 & \color{forestgreen}{\textbf{ENTY}}  & $a$=$1.0$ & {\small \verb!(( |^)(name|name)[^\w]*( |$))!} \\
          Rule 59 & \color{forestgreen}{\textbf{ENTY}}  & $a$=$1.0$ & {\small \verb!( |^)(which|what|what)[^\w]* !} \\
        &  &  & {\small \verb!*([^\s]+ )*(organization|trust|company|company)[^\w]*( |$)!} \\
          Rule 67 & \color{forestgreen}{\textbf{ENTY}}  & $a$=$0.6$ & {\small \verb!( |^)(what|what)[^\w]* (\w+ ){0,1}(is|is)[^\w]* !} \\
        &  &  & {\small \verb!*([^\s]+ )*(surname|address|name|name)[^\w]*( |$)!} \\
    \bottomrule 
    \end{tabular}
    \caption{The clean label in this case is Human, while from the text we understand that the label is Entity. }
    \label{tab:my_label}
\end{table*}

\begin{table*}[!htb]
    \centering
    \begin{tabular}{llll}
    \toprule  
        Text & \multicolumn{3}{l}{What does I.V. stand for ?}  \\
        Clean label & ABBR &&\\
        RAN Teacher & \color{forestgreen}{\textbf{ABBR}} &&\\\midrule
        \textbf{Weak Source} & \textbf{Label} & \textbf{Weight}  & \textbf{Feature}\\
        Student & \color{forestgreen}{\textbf{ABBR}} &  $a$=$1.0$ & $h_i$ (contextualized instance embedding)\\
         Rule 42 & \color{red}{\textbf{DESC}}  & $a$=$0.0$ & {\small \verb!( |^)(explain|describe|what|what)[^\w]*( |$)!} \\
          Rule 54 & \color{red}{\textbf{DESC}} & $a$=$0.1$ & {\small \verb!( |^)(how|what|what)[^\w]*!} \\
      &  &  & {\small \verb!* (\w+ ){0,1}(do|does|does)[^\w]*( |$)!} \\
        Rule 68 & \color{forestgreen}{\textbf{ABBR}} & $a$=$1.0$ & {\small \verb!( |^)(what|what)[^\w]* (\w+ ){0,1}(does|does)[^\w]*!} \\
      &  &  & {\small \verb!* * ([^\s]+ )*(stand for)[^\w]*( |$)!} \\
    \bottomrule 
    \end{tabular}
    \caption{TREC example.}
    \label{tab:my_label}
\end{table*}

\begin{table*}[!htb]
    \centering
    \begin{tabular}{llll}
    \toprule  
        Text & \multicolumn{3}{l}{What year did the Titanic sink ?}  \\
        Clean label & NUM &&\\
        RAN Teacher & \color{forestgreen}{\textbf{NUM}} &&\\\midrule
        \textbf{Weak Source} & \textbf{Label} & \textbf{Weight}  & \textbf{Feature}\\
        Student & \color{forestgreen}{\textbf{NUM}} &  $a$=$1.0$ & $h_i$ (contextualized instance embedding)\\
         Rule 29 & \color{forestgreen}{\textbf{NUM}}  & $a$=$1.0$ & {\small \verb!( |^)(which|what|what)[^\w]* !} \\
           &  &  & {\small \verb!*([^\s]+ )*(time|day|month|hours|minute!} \\
      &  &  & {\small \verb!*|seconds|year|date|date)[^\w]*( |$)!} \\
         Rule 32 & \color{forestgreen}{\textbf{NUM}}  & $a$=$1.0$ & {\small \verb!( |^)(year|year)[^\w]*( |$)!} \\

         Rule 42 & \color{red}{\textbf{DESC}}  & $a$=$0.0$ & {\small \verb!( |^)(explain|describe|what|what)[^\w]*( |$)!} \\
    \bottomrule 
    \end{tabular}
    \caption{TREC example.}
    \label{tab:trec_example_last}
\end{table*}

\end{document}